\documentclass[twoside,11pt]{article}
\usepackage[affil-it]{authblk} 
\usepackage{authblk}       
\usepackage[margin=1in]{geometry}
\usepackage{bm}
\usepackage{microtype}
\usepackage{graphicx}
\usepackage{multirow}
\usepackage{amsmath}
\usepackage{amssymb}
\usepackage{amsfonts}
\usepackage{amsthm}
\usepackage{mathrsfs}
\usepackage[title]{appendix}
\usepackage[svgnames]{xcolor}
\usepackage{textcomp}
\usepackage{manyfoot}
\usepackage{booktabs}
\usepackage{algorithm}
\usepackage{algorithmicx}
\usepackage{algpseudocode}
\usepackage{listings}
\usepackage{mathtools} 
\usepackage{array} 
\usepackage{tikz}
\usepackage{listing}
\usepackage{float}
\usepackage{inconsolata}   
\usepackage{hyperref}
\usepackage[authoryear,round]{natbib}
\newfloat{lstfloat}{htbp}{lop}
\floatname{lstfloat}{Listing}

\usepackage{listofitems} 
\usetikzlibrary{arrows.meta} 
\usetikzlibrary{positioning}
\usepackage[outline]{contour} 
\contourlength{1.4pt}
\tikzset{>=latex} 

\usepackage{xcolor}
\colorlet{outputcolor}{white!80!black}
\colorlet{hiddencolor}{white!20!black}
\colorlet{inputcolor}{white!80!black}
\colorlet{linecolor}{black!40!black}

\newtheorem{theorem}{Theorem}
\newtheorem{lemma}{Lemma}

\newtheorem{property}[theorem]{Property}
\newtheorem{definition}{Definition}

\tikzset{
  >=latex, 
  node/.style={thick,circle,draw=hiddencolor,minimum size=22,inner sep=0.5,outer sep=0.6},
  node in/.style={node,green!20!black,draw=inputcolor!30!black,fill=inputcolor!25},
  node hidden/.style={node,blue!20!black,draw=hiddencolor!30!black,fill=hiddencolor!20},
  node out/.style={node,red!20!black,draw=outputcolor!30!black,fill=outputcolor!20},
  connect/.style={thick,linecolor}, 
  connect arrow/.style={-{Latex[length=4,width=3.5]},thick,linecolor,shorten <=0.5,shorten >=1},
  node 1/.style={node in}, 
  node 2/.style={node hidden},
  node 3/.style={node out}
}
\def\nstyle{int(\lay<\Nnodlen?min(2,\lay):3)} 

\newcolumntype{W}[1]{>{$\displaystyle}w{l}{#1}<{$}}
\newcolumntype{L}{>{\displaystyle}l}

\newcommand{\const}[1]{\ensuremath{\mathrm{#1}}} 
\newcommand{\func}[1]{\ensuremath{\textsf{#1}}} 
\newcommand{\set}[1]{\ensuremath{\{ #1 \} }} 

\newcommand{\Gradient}{\textsf{D}}
\newcommand{\Derivative}[2]{\ensuremath{\frac{\partial #1}{\partial #2}}}
\newcommand{\Partial}[1]{\frac{\partial}{\partial #1}}

\newcommand{\Reals}{\mathbb{R}}

\definecolor[named]{ACMBlue}{cmyk}{1,0.1,0,0.1}
\definecolor[named]{ACMYellow}{cmyk}{0,0.16,1,0}
\definecolor[named]{ACMOrange}{cmyk}{0,0.42,1,0.01}
\definecolor[named]{ACMRed}{cmyk}{0,0.90,0.86,0}
\definecolor[named]{ACMLightBlue}{cmyk}{0.49,0.01,0,0}
\definecolor[named]{ACMGreen}{cmyk}{0.20,0,1,0.19}
\definecolor[named]{ACMPurple}{cmyk}{0.55,1,0,0.15}
\definecolor[named]{ACMDarkBlue}{cmyk}{1,0.58,0,0.21}
\definecolor[named]{ACMTangerine}{cmyk}{0,0.60,0.95,0.01}

\hypersetup{
  colorlinks,
  linkcolor=ACMDarkBlue,  
  citecolor=ACMDarkBlue,
  urlcolor=black,
  filecolor=ACMDarkBlue,
}

\newcommand{\keywords}[1]{\textbf{Keywords:} #1}

\begin{document}

\title{\centering Batch Matrix-form Equations and Implementation \\ of Multilayer Perceptrons}

\author[1]{Wieger Wesselink}
\author[1]{Bram Grooten}
\author[1]{Huub van de Wetering}
\author[1]{Qiao Xiao}
\author[1,2]{Decebal~Constantin Mocanu}

\affil[1]{Eindhoven University of Technology, The Netherlands}
\affil[2]{University of Luxembourg, Luxembourg}
\affil[ ]{}
\affil[ ]{\small\texttt{\{j.w.wesselink,b.grooten,h.v.d.wetering,q.xiao\}@tue.nl, decebal.mocanu@uni.lu}}

\date{}

\maketitle

\begin{abstract}
Multilayer perceptrons (MLPs) remain fundamental to modern deep learning, yet their algorithmic details are rarely presented in complete, explicit \emph{batch matrix-form}. Rather, most references express gradients per sample or rely on automatic differentiation. Although automatic differentiation can achieve equally high computational efficiency, the usage of batch matrix-form makes the computational structure explicit, which is essential for transparent, systematic analysis, and optimization in settings such as sparse neural networks. This paper fills that gap by providing a mathematically rigorous and implementation-ready specification of MLPs in batch matrix-form. We derive forward and backward equations for all standard and advanced layers, including batch normalization and softmax, and validate all equations using the symbolic mathematics library SymPy. From these specifications, we construct uniform reference implementations in NumPy, PyTorch, JAX, TensorFlow, and a high-performance C++ backend optimized for sparse operations.
Our main contributions are:
(1) a complete derivation of batch matrix-form backpropagation for MLPs,
(2) symbolic validation of all gradient equations,
(3) uniform Python and C++ reference implementations grounded in a small set of matrix primitives, and
(4) demonstration of how explicit formulations enable efficient sparse computation.
Together, these results establish a validated, extensible foundation for understanding, teaching, and researching neural network algorithms.
\end{abstract}

\begin{center}
\parbox{0.88\textwidth}{ 
  \keywords{
    batch matrix-form backpropagation, 
    multilayer perceptrons, 
    symbolic equation validation, 
    sparse neural networks, 
    reference implementations
  }
}
\end{center}

\section{Introduction}
The inner workings of multilayer perceptrons (MLPs) are often treated as implementation details hidden within large frameworks. As a result, the precise mathematical structure of forward and backward computations, especially in batch matrix-form, is rarely documented explicitly. This creates a gap between conceptual understanding and practical implementation, limiting both educational use and the development of optimized or sparse neural architectures.

Neural networks are widely implemented in frameworks such as PyTorch \citep{DBLP:conf/nips/PaszkeGMLBCKLGA19}, JAX \citep{jax2018github}, and TensorFlow \citep{DBLP:conf/osdi/AbadiBCCDDDGIIK16}. While these frameworks perform efficiently for end users, they abstract away the underlying algorithms. This lack of transparency can indeed hinder students who aim to understand neural network mechanics or researchers who wish to modify and experiment with them. In particular, backpropagation equations are usually implicit, hidden behind automatic differentiation tools \citep{DBLP:journals/jmlr/BaydinPRS17}, which can limit applications requiring explicit matrix computations. Explicit derivations of batch matrix-form backpropagation equations are rarely provided in the literature and, to our knowledge, are absent for advanced layers such as batch normalization and softmax.
To address this, the Nerva project \citep{Wesselink_Nerva_library} provides a collection of C++ and Python libraries that implement multilayer perceptrons (MLPs) using explicit batch matrix-form equations and readable reference implementations.

Sparse neural networks provide a clear example of why explicit backpropagation equations are useful. In \citet{wesselink2024nervatrulysparseimplementation}, we evaluate the performance of truly sparse layers. Since the support for sparse tensors in Python frameworks is still limited or experimental, we opt for a native C++ implementation using the Intel MKL Library for efficient sparse matrix operations. By employing explicit backpropagation equations, we not only circumvent the insufficient support for automatic differentiation in sparse computations, but also gain additional advantages: precise identification of the sparse and dense operations that dominate performance, and systematic exploration of optimizations.

The main focus of this paper remains the practical implementation of MLPs. We provide precise mathematical specifications of MLPs in explicit batch matrix-form, including complete backpropagation equations for standard and advanced layers, and produce straightforward, readable implementations that closely reflect these specifications. We include full matrix-form equations for commonly used layers, activation functions, and loss functions, along with derivations. Correctness is crucial; we therefore rely on the symbolic Python library SymPy \citep{10.7717/peerj-cs.103}, which enables early validation of equations and precise localization of errors in formulas. This goes beyond standard numerical gradient checking, which only indicates that an error exists without identifying its source.

We express the execution of MLPs using a carefully selected set of matrix operations. Table \ref{table:matrix-operations} provides an overview of these operations, and Table \ref{table:matrix-operations-implementation} shows their realizations in Python frameworks such as NumPy \citep{harris2020array}, PyTorch, JAX, and TensorFlow, as well as in C++ using Eigen \citep{eigenweb}. These tables serve as a bridge from mathematical equations to code, yielding implementations that are uniform, maintainable, and readily extensible to other frameworks. All implementations presented in this paper are part of the Nerva project \citep{Wesselink_Nerva_library} and are derived from explicit batch matrix-form equations, expressed in the set of matrix operations in Table~\ref{table:matrix-operations}. The C++ backends target high-performance computation with support for truly sparse networks, while the Python backends emphasize clarity and readability to support education and experimentation. Together, they provide complementary perspectives: performance-oriented implementations in C++ and transparent reference implementations in Python.
Summarizing, this paper makes the following contributions:
\begin{enumerate}
    \item We present a complete derivation of batch matrix-form backpropagation for multilayer perceptrons (MLPs), covering both standard and advanced layers such as batch normalization and softmax.
    \item We symbolically validate all gradient equations using the SymPy library, providing mathematical assurance beyond numerical gradient checking.
    \item We provide uniform reference implementations across major Python frameworks (Num\-Py, PyTorch, JAX, TensorFlow) and a high-performance C++ backend, all grounded in the same small set of matrix primitives.
    \item We demonstrate how explicit matrix-form formulations make the computational structure of MLPs visible, enabling systematic optimization, exploration of sparsity, and support for specialized architectures.
\end{enumerate}

The remainder of this paper is structured as follows: Section \ref{section:mathematical-background} provides mathematical background, Section \ref{section:multilayer-perceptrons} introduces MLPs and their training via stochastic gradient descent. Section \ref{section:sparse-neural-networks} covers sparse neural networks and their implementation in Nerva, while the design and implementation details are described in Section \ref{section:design-implementation-mlps}. Section \ref{section:experimental-results} presents experimental results and we conclude the paper in Section \ref{section:conclusions}. Appendices include layer equations (\ref{appendix:layers}), matrix operation implementations (\ref{appendix:matrix_operations}), activation functions (\ref{appendix:activation-functions}), loss functions (\ref{appendix:loss-functions}), weight initialization (\ref{appendix:weight-initialization}), optimization functions (\ref{appendix:optimization}), and learning rate schedulers (\ref{appendix:learning-rate-schedulers}), with derivations provided where relevant.


\section{Mathematical Background} \label{section:mathematical-background}
In this section we provide notational conventions, we provide definitions for the gradient and the Jacobian, and a comprehensive table with matrix operations that are necessary for the execution of multilayer perceptrons.

In the rest of this paper we use the following notations. Vectors are denoted using lowercase symbols $x, y, z$, and matrices using uppercase symbols $X, Y, Z$. The columns of a matrix $X \in \Reals^{m \times n}$ are denoted as $x^1, \ldots, x^n$, while the rows are denoted as $x_1, \ldots, x_m$. Occasionally the subscript notation $x_i$ is also used to denote the $i^\text{th}$ element of vector $x$, but this will be clear from context.
To distinguish between row and column vectors, we denote their domains as $\Reals^{1 \times n}$ and $\Reals^{m \times 1}$, respectively. We occasionally use the dot symbol~$\cdot$ to denote matrix multiplication; we never use it for the vector dot product.

\begin{figure*}[bth]
    \centering
    \includegraphics[width=\textwidth]{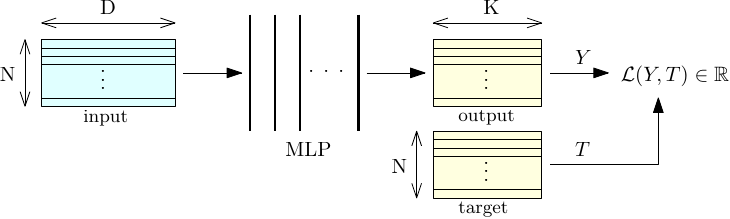}
    \caption{Graphical representation of a multilayer perceptron. The input is a mini-batch, represented by an $\const{N} \times \const{D}$ matrix consisting of $\const{N}$ examples with $\const{D}$ features. In the feedforward pass this input is passed through an MLP with a series of layers, represented by vertical bars. This results in an $\const{N} \times \const{K}$ output matrix $Y$. From the output $Y$ and the expected output $T$ the loss $\mathcal{L}(Y, T)$ is computed.}
    \label{fig:mlp}
\end{figure*}

Having established these general notations, we now apply them in the context of neural networks. Throughout the paper, $x, X$ denote inputs, $y, Y$ outputs, $z, Z$ intermediate values, and $t, T$ targets. The number of inputs of a layer is denoted by $\const{D}$ and the number of outputs by $\const{K}$. $\const{N}$ indicates the number of samples in a mini-batch.
We follow the convention of the major neural network frameworks where data is stored using a row layout. This means that by default $x$, $y$ and $z$ are considered to be row vectors, and that each row of an input matrix $X$ represents an example of a dataset. For example, if $X \in \Reals^{\const{N} \times \const{D}}$, then $x_i \in \Reals^{1 \times \const{D}}$ represents the $i$-th example.
Figure \ref{fig:mlp} provides a graphical representation of a multilayer perceptron, illustrating the notation introduced above.

We express core concepts used throughout our work in the following definitions.
\begin{definition}[Gradient]
Let $f: \Reals^{m \times n} \rightarrow \Reals$ be a function with input $X$ that has elements $x_{ij}$, with $m, n \in \mathbb{N}^{+}$. Then the gradient $\nabla_X f$ is defined as
\begin{equation}
\nabla_X f(X) =
\begin{bmatrix}
 \dfrac{\partial f(X)}{\partial x_{11}} & \cdots & \dfrac {\partial f(X)}{\partial x_{1n}}
   \\
   \vdots & \ddots & \vdots \\
   \dfrac {\partial f(X)}{\partial x_{m1}} & \cdots & \dfrac {\partial f(X)}{\partial x_{mn}}
\end{bmatrix}.
\end{equation}
\end{definition}

To streamline later derivations, we introduce a shorthand notation for the gradient of the loss. 
\begin{definition}[Gradient of the loss function] \label{def:gradient}
For neural networks with output $Y \in \Reals^{\const{N} \times \const{K}}$ and target $T \in \Reals^{\const{N} \times \const{K}}$, let $\mathcal{L}: \Reals^{\const{N} \times \const{K}} \times \Reals^{\const{N} \times \const{K}} \rightarrow \Reals$ be a fixed loss function. The gradient of the loss with respect to the output is written as
\begin{equation}
\Gradient Y = \nabla_Y \mathcal{L}(Y, T).
\end{equation}
More generally, if $Y$ depends on a parameter $Z$ (i.e., $Y = Y(Z)$), we define
\begin{equation}
\Gradient Z = \nabla_Z \mathcal{L}(Y(Z), T).  
\end{equation}
\end{definition}

\begin{definition}[Jacobian]
Let $f: \Reals^n \rightarrow \Reals^m$ be a function with input $x \in \Reals^n$. Then the Jacobian $\frac{\partial f}{\partial x}$ is defined as follows.
\begin{equation}
\frac{\partial f}{\partial x}(x) =
\begin{bmatrix}
 \dfrac{\partial f_{1}(x)}{\partial x_{1}} & \cdots & \dfrac {\partial f_{1}(x)}{\partial x_{n}}
   \\
   \vdots & \ddots & \vdots \\
   \dfrac {\partial f_{m}(x)}{\partial x_{1}} & \cdots & \dfrac {\partial f_{m}(x)}{\partial x_{n}}
\end{bmatrix}.
\end{equation}
We adopt the convention that the Jacobian does not depend on whether $x$ and $f(x)$ are represented as row or column vectors. This choice is consistent with the symbolic Python library SymPy, which we use for validation, although other conventions exist in the literature.
\end{definition}

The execution of MLPs depends on a small number of matrix operations. In Table \ref{table:matrix-operations} on page~\pageref{table:matrix-operations} we provide a mathematical notation, a code representation, and a definition for the most important operations. These operations are used in the implementation of activation functions, loss functions, and the feedforward and backpropagation equations of neural network layers. Basic operations like matrix assignment, matrix indexing and matrix dimensions are omitted, to keep the resulting code familiar to users of the respective Python frameworks. There is some redundancy in the table, to allow for more efficient, or numerically stable implementations.
We refrain from using broadcasting notation, as commonly found in frameworks like NumPy, where arrays of different dimensions can be added and use the standard notation of matrix calculus instead.
This approach with explicit dimensions is needed for validating the correctness using SymPy.
For instance, to compute the sum of elements in a column vector $x \in \Reals^{n \times 1}$, we express it as the dot product $1_n^\top \cdot x$, with $1_n = (1, \ldots, 1)^\top$ a column vector of ones.

\begin{table*}[thbp]
\centering
\caption{An overview of matrix operations that are needed for the implementation of the class of MLPs described in this paper. We assume that $X,Y \in \Reals^{m \times n}$ and $Z \in \Reals^{n \times k}$, where $k, m, n \in \mathbb{N}^{+}$. Wherever possible, we use $m$
to denote the number of rows and $n$ to denote the number of columns of a matrix or vector.
The function $\sigma$ is the sigmoid function as defined in Appendix \ref{appendix:activation-functions}.
In the second column we use the unified Nerva API notation, which is implemented consistently across the Python and C++ backends, ensuring a one-to-one correspondence between equations and code.}
\label{table:matrix-operations}
{
\small
\begin{align*}
\toprule
    \textsc{operation} &\quad \textsc{Nerva API} & \textsc{definition}
    \\
    \midrule
    0_{m} &\quad \texttt{zeros(m)} & \text{$m \times 1$ column vector with elements equal to 0}
    \\
    0_{mn} &\quad \texttt{zeros(m, n)} & \text{$m \times n$ matrix with elements equal to 0}
    \\
    1_{m} &\quad \texttt{ones(m)} & \text{$m \times 1$ column vector with elements equal to 1}
    \\
    1_{mn} &\quad \texttt{ones(m, n)} & \text{$m \times n$ matrix with elements equal to 1} 
    \\
    \mathbb{I}_n &\quad \texttt{identity(n)} & \text{$n \times n$ identity matrix} 
    \\
    X^\top &\quad \texttt{X.T} & \text{transposition}
    \\
    cX &\quad \texttt{c * X} & \text{scalar multiplication, $c \in \Reals$}
    \\
    X + Y &\quad \texttt{X + Y} & \text{addition}
    \\
    X - Y &\quad \texttt{X - Y} & \text{subtraction}
    \\
    X \cdot Z &\quad \texttt{X @ Z} \text{ or } \texttt{X * Z} 
    & \text{matrix multiplication, also denoted as $XZ$}
    \\
    x^\top y \text{ or } x y^\top &\quad \texttt{dot(x,y)} & \text{dot product, } x,y \in \Reals^{m \times 1} \text{ or } x,y \in \Reals^{1 \times n}
    \\
    X \odot Y &\quad \texttt{hadamard(X,Y)} & \text{element-wise product of $X$ and $Y$}
    \\
    \func{diag}(X) &\quad \texttt{diag(X)} & \text{column vector that contains the diagonal of $X$} 
    \\
    \func{Diag}(x) &\quad \texttt{Diag(x)} &\quad \text{diagonal matrix with $x$ as diagonal, } x \in \Reals^{1 \times n} \text{ or } x \in \Reals^{m \times 1} 
    \\
    1_m^\top \cdot X \cdot 1_n &\quad \texttt{elements\_sum(X)} & \text{sum of the elements of $X$}
    \\
    x \cdot 1_n^\top &\quad \texttt{column\_repeat(x, n)} & \text{$n$ copies of column vector $x \in \Reals^{m \times 1}$} 
    \\
    1_m \cdot x &\quad \texttt{row\_repeat(x, m)} & \text{$m$ copies of row vector $x \in \Reals^{1 \times n}$}
    \\
    1_m^\top \cdot X &\quad \texttt{columns\_sum(X)} & \text{$1 \times n$ row vector with sums of the columns of $X$}
    \\
    X \cdot 1_n &\quad \texttt{rows\_sum(X)} & \text{$m \times 1$ column vector with sums of the rows of $X$} 
    \\
    \max(X)_\text{col} &\quad \texttt{columns\_max(X)} & \text{$1 \times n$ row vector with maximum values of the columns of $X$}
    \\
    \max(X)_\text{row} &\quad \texttt{rows\_max(X)} & \text{$m \times 1$ column vector with maximum values of the rows of $X$}
    \\
    (1_m^\top \cdot X) / n &\quad \texttt{columns\_mean(X)} & \text{$1 \times n$ row vector with mean values of the columns of $X$}
    \\
    (X \cdot 1_n) / m &\quad \texttt{rows\_mean(X)} & \text{$m \times 1$ column vector with mean values of the rows of $X$}
    \\
    f(X) &\quad \texttt{apply(f, X)} & \text{element-wise application of $f: \Reals \rightarrow \Reals$ to $X$} 
    \\
    e^X &\quad \texttt{exp(X)} & \text{element-wise application of $f: x \rightarrow e^x$ to $X$} 
    \\
    \log(X) &\quad \texttt{log(X)} & \text{element-wise application of the natural logarithm $f: x \rightarrow \ln(x)$ to $X$}
    \\
    1 / X &\quad \texttt{reciprocal(X)} & \text{element-wise application of $f: x \rightarrow 1/x$ to $X$}
    \\
    \sqrt{X} &\quad \texttt{sqrt(X)} & \text{element-wise application of $f: x \rightarrow \sqrt{x}$ to $X$}
    \\ X^{-1/2} &\quad \texttt{inv\_sqrt}(X) & \text{element-wise application of $f: x \rightarrow x^{-1/2}$ to $X$} 
    \\
    \log(\sigma(X)) &\quad \texttt{log\_sigmoid}(X) & \text{element-wise application of $f: x \rightarrow \log(\sigma(x))$ to $X$},
    \\
    \bottomrule
\end{align*}
}
\end{table*}


\section{Multilayer Perceptrons} 
\label{section:multilayer-perceptrons}
\begin{figure}[tbh]
\centering
\begin{tikzpicture}[x=1.7cm,y=1.2cm,baseline=(N1-1)]
  \message{^^JNeural network, shifted}
  \readlist\Nnod{3,4,4,4,3} 
  \readlist\Nstr{D,m,m,m,K} 
  \readlist\Cstr{\strut x,a^{(\prev)},a^{(\prev)},a^{(\prev)},y} 
  \def\yshift{0.5} 
  
  \message{^^J  Layer}
  \foreachitem \N \in \Nnod{ 
    \def\lay{\Ncnt} 
    \pgfmathsetmacro\prev{int(\Ncnt-1)} 
    \message{\lay,}
    \foreach \i [evaluate={\c=int(\i==\N); \y=\N/2-\i-\c*\yshift;
                 \index=(\i<\N?int(\i):"\Nstr[\lay]");
                 \x=\lay; \n=\nstyle;}] in {1,...,\N}{ 
      \ifnum\lay=1 
        \node[node \n] (N\lay-\i) at (\x,\y) {$\Cstr[\lay]_{\index}$};
      \else 
        \ifnum\lay=\Nnodlen 
          \node[node \n] (N\lay-\i) at (\x,\y) {$\Cstr[\lay]_{\index}$};
        \else 
          \node[node \n] (N\lay-\i) at (\x,\y) {};
        \fi
      \fi
      
      \ifnum\lay>1 
        \foreach \j in {1,...,\Nnod[\prev]}{ 
          \draw[connect,white,line width=1.2] (N\prev-\j) -- (N\lay-\i);
          \draw[connect arrow] (N\prev-\j) -- (N\lay-\i);
        }
      \fi 
      
    }
    \path (N\lay-\N) --++ (0,1+\yshift) node[midway,scale=1.5] {$\vdots$};
  }
  
  \node[above=0,align=center,inputcolor!60!black] at (N1-1.90) {input\\[-0.2em]layer};
  \node[above=0,align=center,hiddencolor!60!black] at (N3-1.90) {hidden layers};
  \node[above=0,align=center,outputcolor!60!black] at (N\Nnodlen-1.90) {output\\[-0.2em]layer}; 
\end{tikzpicture}
    \caption{A multilayer perceptron with three hidden layers. The input vector $x$ has $\const{D}$ elements, and the output vector $y$ has $\const{K}$ elements. Neurons in each layer are represented by circles, and connections between neurons are indicated by arrows. The weights associated with the connections form the parameters of the linear layers.}
    \label{fig:multilayer-perceptron}
\end{figure}
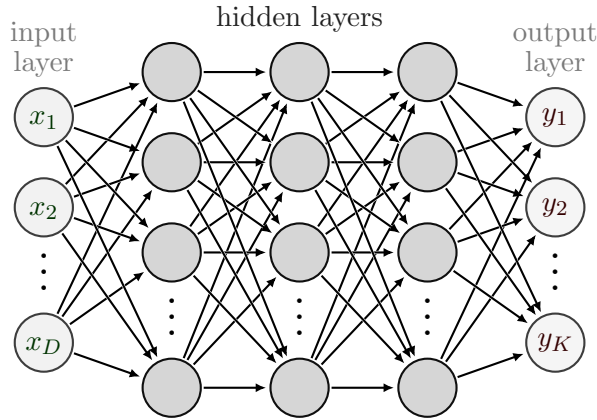

A multilayer perceptron is an artificial neural network comprised of layers, each containing neurons, which are the basic processing units, see Fig. \ref{fig:multilayer-perceptron}.
The network typically consists of three types of layers: an input layer, a number of hidden layers, and an output layer. Neurons in the input layer represent features of the input data, while neurons in subsequent layers capture increasingly complex patterns.
At the core of MLPs lies the concept of linear layers. In a linear layer, the input row vector $x$ undergoes a linear transformation represented by  $x W^\top + b$, where $W$ is a weight matrix\footnote{Matrix $W$ is transposed to be consistent with existing frameworks.} and $b$ is a bias vector.
In matrix-form, where each row of a batch $X$ is an input, this turns into
\begin{equation} \label{eq:linear-layer-feedforward}
    Y = X \cdot W^{\top} + 1_\const{N} \cdot b.
\end{equation}
Since linear transformations are limited in their expressive power, activation functions \citep{DBLP:journals/ijon/DubeySC22} can be applied to the output of linear layers to introduce non-linearity to the model. Common activation functions include the sigmoid, hyperbolic tangent (tanh), and rectified linear unit (ReLU). These functions allow the network to learn complex, non-linear patterns in the data.
Beyond linear and activation layers, additional layers like dropout and batch normalization contribute to the performance and stability of MLPs. 
A loss function is used to quantify the disparity between the predicted outputs and the true targets. Common loss functions include mean squared error for regression tasks and cross-entropy loss for classification tasks.
The term \textit{feedforward} describes the process of passing data through the network from input to output. \textit{Backpropagation} \citep{kelley1960gradient} is the mechanism used to iteratively adjust the weights and biases, making the neural network ``learn''. This is done based on the gradients with respect to the loss function that are computed during a backward pass. This iterative process refines the model's ability to make accurate predictions.
This adjustment of the network parameters is called optimization. Gradient Descent, a popular optimization method, minimizes the loss function by iteratively moving in the direction of steepest descent in the parameter space. Modern variations, like Adam \citep{DBLP:journals/corr/KingmaB14}, incorporate adaptive learning rates, enhancing convergence speed and stability.
In summary, MLPs constitute a powerful framework for modeling complex relationships within data. 
MLPs are universal function approximators, meaning they can theoretically learn and approximate any continuous function. An extensive treatment of MLPs (without matrix-form equations) can be found in
\citep{zhang2024dive}.

Mathematically, a neural network can be viewed as a function $f_\Theta: \Reals^\const{D} \rightarrow \Reals^\const{K}$ that depends on a set of parameters $\Theta$ (including the weights and biases of linear layers), and that maps an input vector $x \in \Reals^\const{D}$ to an output vector $y \in \Reals^\const{K}$. We consider the supervised learning regime, where we have a dataset of inputs $\set{ x_i }_{1 \leq i \leq \const{N}}$, and a corresponding set of targets $\set{t_i}_{1 \leq i \leq \const{N}}$. The output is denoted as $y_i = f_\Theta(x_i)$. The goal of training a neural network is to find values for $\Theta$ that minimize the loss $\sum_{i=1}^{\const{N}} \mathcal{L}(y_i, t_i)$,
where $\mathcal{L}$ is a loss function that measures a distance between its arguments $y$ and $t$. In practical applications, a neural network can handle multiple inputs at once. This is done by taking a matrix $X \in \Reals^{\const{N} \times \const{D}}$ as input (a mini-batch) together with a matrix $T \in \Reals^{\const{N} \times \const{K}}$, such that the rows of $X$ and $T$ are the inputs with their corresponding targets.  We can lift the function $f_\Theta$ to multiple inputs using the matrix-form notation $Y = f_\Theta(X)$, where $y_i = f_\Theta(x_i)$ for $1 \leq i \leq \const{N}$. The function $\mathcal{L}$ is expanded to multiple inputs using
\[
  \mathcal{L}(Y, T) = \sum_{i=1}^\const{N} \mathcal{L}(y_i, t_i).
\]
When using mini-batches as input, the training of a neural network can be expressed using matrix-matrix products instead of matrix-vector products. This generally increases the performance significantly,
since there are specialized libraries available that apply massive parallelism and hardware optimizations like vectorization. For this reason we will provide specifications of neural networks that handle inputs in matrix-form.

\subsection{Batch normalization}
Batch normalization \citep{DBLP:conf/icml/IoffeS15} can be used to normalize the inputs of neural network layers by adjusting the mean and variance of the data. This can help to speed up and stabilize training by mitigating issues such as vanishing or exploding gradients.
We will slightly rewrite the formulation of \citet{DBLP:conf/icml/IoffeS15}, and convert it into matrix-form. Note that the input and output size of a batch normalization layer are the same, i.e., $\const{D} = \const{K}$. Let $X \in \Reals^{\const{N} \times \const{D}}$ be an input batch.
Each row of $X$ corresponds with an example, and each column with a feature. Let $\mu = (1_\const{N}^\top \cdot X) / \const{N}$, i.e., the row vector $\mu \in \Reals^{1 \times \const{D}}$ contains the averages of the columns in the batch. For a given example $x_i$, we write $r_i = x_i - \mu \in \Reals^{1 \times \const{D}}$, which represents the centered feature vector of that example. For a given feature $j$, we define $\sigma_j^2 = ((r^j)^\top \cdot r^j) / \const{N}$, which is the variance of the $j$-th feature across the batch.
The variance vector $\Sigma \in \Reals^{1 \times \const{D}}$ that contains $\sigma_j^2$ as its elements can then be formulated as $\Sigma = \func{diag}(R^\top R)^\top/\const{N}$, where $R$ is the $\const{N} \times \const{D}$ matrix with rows $r_1, \ldots, r_\const{N}$.
The first part of batch normalization, which standardizes the input $X$ to a matrix $Z$ with mean 0 and standard deviation 1, is then given by the equations
\begin{align}
\begin{split}
R &= X - \frac{1_\const{N} \cdot 1_\const{N}^\top}{\const{N}} \cdot X
\\
\Sigma &= \frac{1}{\const{N}} \cdot \func{diag}(R^\top R)^\top
\\
Z &= (1_\const{N} \cdot \Sigma^{-\frac{1}{2}}) \odot R.
\end{split}
\end{align}
The second part of batch normalization consists of a scale and a shift operation that transforms the rows $z_1, \ldots, z_\const{N}$ of $Z$ using $y_i = \gamma \odot z_i + \beta$, with $\beta, \gamma \in \Reals^{1 \times \const{D}}$. In matrix-form this becomes
\begin{align}
Y &= (1_\const{N} \cdot \gamma) \odot Z + 1_\const{N} \cdot \beta.
\end{align}
The parameters $\beta$ and $\gamma$ are learnable parameters, meaning that they are periodically updated using an optimizer function in order to minimize the loss. Note that each column of $R$ sums to zero, which is a direct consequence of subtracting the average $\mu$ from the rows of $X$. Since $Z$ scales the columns of $R$ with a constant factor, the matrix $Z$ has the same property. This property will be exploited in the derivation of the backpropagation equations in Appendix \ref{appendix:layers}.

\subsection{Dropout}

Dropout \citep{DBLP:journals/jmlr/SrivastavaHKSS14} is a technique that can be used to prevent \emph{overfitting} of a neural network. Multiple variants of dropout exists. In this section we discuss the DropConnect variant \citep{DBLP:conf/icml/WanZZLF13}. During training of the network it randomly replaces a fraction $0 < p < 1$ of the weights of a linear layer (i.e., the elements of the matrix $W$) by zero, for example at the start of each epoch of training. An easy way to implement this is to define a matrix $R$
(not to be confused with the $R$ used in batch normalization)
with the same dimensions as the weight matrix $W$ and that contains the value 0 on all positions that are dropped, and the value 1 on all other positions. The effect of dropping the weights can then be achieved by using the weight matrix $W' = W \odot R$ instead of $W$. 
However, doing this has an unwanted side effect: it will cause the sum of the absolute values of the weights to decrease with a factor $1 - p$ on average. To compensate for this, the non-zero values of $R$ should get the value $1 / (1 - p)$ instead of 1.
The feedforward equation of dropout that we use is therefore
\begin{align}
Y &= X (W \odot R)^\top + 1_\const{N} \cdot b.
\end{align}
Note that this variation of dropout is applied to the connections between the neurons. It is also possible to apply dropout to the neurons of a layer.

\subsection{Training a multilayer perceptron}
The first step of training an MLP is to choose an architecture for the network that is suitable for a given task. In case of an MLP this comes down to selecting a sequence of layers, their sizes and activation functions, and  a loss function. How to do this is outside the scope of this paper. However, in the appendices we provide detailed descriptions of several layer types (\ref{appendix:layers}), activation functions (\ref{appendix:activation-functions}) and loss functions (\ref{appendix:loss-functions}).
The next step is to initialize the parameters of the layers, in particular the weights and biases of the linear layers. This too can have a large impact on performance \citep{DBLP:journals/corr/abs-1805-08266}. In Appendix \ref{appendix:weight-initialization} we give a few common weight initialization strategies.
Training an MLP is usually done with (a variant of) stochastic gradient descent and consists of three steps that are performed repeatedly:
\begin{enumerate}
    \item (feedforward) Given an input batch $X$ and the neural network parameters $\Theta$, compute the output $Y$.
    \item (backpropagation) Given outputs $Y$ corresponding to inputs $X$ and expected outputs $T$, compute the gradient of the loss $\Gradient Y$. Then from $Y$ and $\Gradient Y$, compute the gradient $\Gradient \Theta$ of the parameters $\Theta$.
    \item (optimization) Given the gradient $\Gradient \Theta$, update the parameters $\Theta$.
\end{enumerate}
\begin{figure}[t]
    \centering
    \begin{tikzpicture}[>=Stealth, node distance=1cm]
    
    \node (X) at (1,0) {$X$};
    \node (Theta) [right=of X] {$\Theta$};
    \node (DX) [below=of X] {$\const{D}X$};
    \node (DTheta) [below=of Theta] {$\const{D}\Theta$};
    
    \node (layer) at (1.75,-0.75) [draw, dashed, rounded corners=5mm, rectangle, minimum width=3cm, minimum height=2.5cm] {};
    \node (layerlabel) [font=\small, above=-1.55cm of layer] {\textsf{Layer}};
    
    \node (Y) [draw, circle, minimum size=0.9cm, right=3.5cm of Theta] {$Y$};
    
    \node (DY) [draw, circle, minimum size=0.9cm, below=0.6cm of Y] {$\const{D}Y$};
    
    \draw[->] (3.25, 0) -- (6.32, 0) node[font=\small, pos=0.2, above=0.2cm of Theta, right] {\textsf{feedforward}};
    \draw[->] (6.3, -1.55) -- (3.25, -1.55) node[font=\small, pos=0.95, above=0.2cm of DTheta, right] {\textsf{backpropagation}};
    \draw[-] (6.75, -0.45) -- (6.75, -0.75);
    \draw[-] (6.75, -0.75) -- (6.0, -0.75);
    \draw[-] (6.0, -0.75) -- (6.0, -1.55);
    
    \end{tikzpicture}

    \caption{Data flow of a layer. A layer stores the input $X$ and the layer parameters $\theta$ (e.g., the weight matrix $W$ and the bias vector $b$ in case of a linear layer), and the corresponding gradients $\Gradient X$ and $\Gradient \theta$ with respect to the loss function. In the feedforward step the output $Y = f_\theta(X)$ is calculated and forwarded to the next layer. In the backpropagation step the output $Y$ and its gradient $\Gradient Y$ are obtained from the next layer and used to calculate the gradients $\Gradient X$ and $\Gradient \theta$. $X$ and $\Gradient X$ are then passed back to the previous layer. In the optimization step the gradient $\Gradient \theta$ is used to update the value $\theta$.
    }
    \label{fig:layers}
\end{figure}
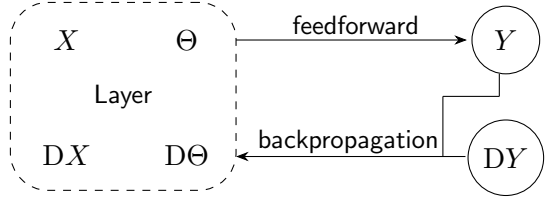
These steps are performed for each input batch $X$ with target $T$ of a dataset, and this process is repeated a given number of times (the number of epochs). For the optimization step a learning rate \citep{DBLP:conf/bigdataconf/WuLBCIPWYZ19} is used to determine
the size of the steps taken. It controls how much the model's weights should be adjusted with respect to the gradient of the loss function. The learning rate is often decreased during training, in Appendix \ref{appendix:learning-rate-schedulers} we provide an overview of some learning rate schedulers.

\section{Sparse Neural Networks} \label{section:sparse-neural-networks}
This section presents a case study for assessing the performance of truly sparse neural networks, using the C++ backend of the Nerva libraries \citep{Wesselink_Nerva_library}. The study reported in \citep{wesselink2024nervatrulysparseimplementation} serves as an example of the approach advocated in this paper: implementing neural networks with explicit backpropagation equations in batch matrix-form and clearly defined matrix operations.

The practical implementation of sparse neural networks remains a significant challenge. While major frameworks are actively extending sparse tensor support, limitations in operator coverage, efficiency, and autograd capabilities persist. Researchers often resort to workarounds such as binary masks applied to dense tensors \citep{liu2023lessonslearnednewsparseland}, which do not deliver the true computational and memory savings of sparsity. As a result, evaluating the performance of truly sparse networks within these ecosystems is difficult.

The Nerva backend forgoes automatic differentiation not only because Eigen does not provide native autograd support, but also because existing C++ autodiff libraries are not well-optimized for sparse parameters. Instead, it uses manually derived backpropagation equations in batch matrix-form.
Using these explicit equations has two important benefits:
\begin{enumerate}
\item \textbf{Correctness:} By using validated equations and expressing their implementation in a small set of well-defined matrix operations, the correctness of the implementation follows directly from the validated equations.
\item \textbf{Profiling and optimization:} Each equation corresponds to one or more matrix operations, enabling fine-grained timing and analysis of sub-steps (e.g., in batch normalization) and core sparse/dense products.
\end{enumerate}
Fine-grained profiling reveals that the most critical and performance-sensitive operations in sparse networks are:
\begin{align*}
A &= S B &\hspace{-2cm} \text{(sparse-dense product, feedforward)} \\
A &= S^\top B &\hspace{-2cm} \text{(sparse-dense product, backward pass)} \\
S &= A B^\top &\hspace{-2cm} \text{(dense-dense product, backward pass)}
\end{align*}
where $A$ and $B$ are dense matrices, and $S$ and $T$ are sparse matrices. Among these, the product $S = A B^\top$ is particularly challenging. Because $S$ is sparse, only a small subset of values from the dense product $A B^\top$ are retained. In practice, this operation often dominates backpropagation in linear layers, consuming over half of the backpropagation time in experiments such as training on CIFAR-10 at 95\% sparsity. The right-hand side is a dense matrix product, but only a fraction of its entries are actually used to update the sparse weight matrix. Explicitly forming and storing $AB^\top$ is therefore computationally expensive and negates the memory advantages of sparsity. Several approaches can be used to compute this product directly in sparse form, and we experimented with a number of them in the Nerva libraries.
Additional performance issues appeared in various parts of the implementation. Dense products were sometimes not forwarded by Eigen’s expression trees to MKL, resulting in slower execution, and some MKL sparse kernels exhibited poor efficiency for large matrices. The advantage of expressing the implementation through explicit matrix-form equations is that such bottlenecks become straightforward to detect and analyze.

Empirical results confirm the practical benefits of this approach. While dense MLPs implemented with Nerva perform on par with PyTorch, the advantage becomes clear in sparse settings: PyTorch's masking strategy maintains nearly constant runtime across sparsity levels, whereas Nerva's truly sparse implementation accelerates as sparsity increases. At 99\% sparsity on CIFAR-10, Nerva is about four times faster and uses dramatically less memory, enabling training of networks too large for dense frameworks. These results illustrate how explicit matrix-form equations combined with an efficient sparse backend, as promoted in this paper, provide a transparent path to studying, optimizing, and scaling truly sparse neural networks beyond the limits of dense frameworks.


\section{Design and implementation of multilayer perceptrons} \label{section:design-implementation-mlps}
In this section we explain our implementation. We have chosen for an object oriented design in which
layers are represented by classes. The main algorithmic structure is presented using our NumPy framework, but it is similar to all others. 

Both an MLP and its layers support a feedforward computation, a backpropagation and an optimization step, see Listing \ref{listing:mlp-implementation}.
\begin{lstfloat}
\begin{lstlisting}
class Layer(object):
    def feedforward(self, X: Matrix) -> Matrix:
    def backpropagate(self, Y: Matrix, DY: Matrix):
    def optimize(self, eta: float):

class MultilayerPerceptron(object):
    def feedforward(self, X: Matrix) -> Matrix:
        for layer in self.layers:
            X = layer.feedforward(X)
        return X

    def backpropagate(self, Y: Matrix, DY: Matrix):
        for layer in reversed(self.layers):
            layer.backpropagate(Y, DY)
            Y, DY = layer.X, layer.DX

    def optimize(self, eta: float):
        for layer in self.layers:
            layer.optimize(eta)
\end{lstlisting}
\caption{Multilayer perceptron implementation. Only the fundamental methods needed for the execution are given. For completeness the interface of the layer classes is added, which has the same structure. The attributes of
a layer are the input \texttt{X} and its gradient \texttt{DX}, and a function pointer \texttt{optimizer}
that takes care of the optimization step. A multilayer perceptron has only one attribute, \texttt{layers},
for storing the list of layers.
}
\label{listing:mlp-implementation}
\end{lstfloat}
Each layer of an MLP depends on a number of parameters $\theta$.
In a layer we store the input $X$, the parameters $\theta$, their gradients $\Gradient X$ and $\Gradient \theta$, and a function object for the optimization step. During the feedforward step the output $Y$ is calculated and stored as input in the next layer. During the backpropagation step the gradients $\Gradient X$ and $\Gradient \theta$ are computed from the output $Y$ and its gradient $\Gradient Y$, that are obtained from the next layer. This is depicted in Fig. \ref{fig:layers}.
Listing \ref{listing:stochastic-gradient-descent} shows how stochastic gradient descent can be implemented using our NumPy framework. Here \texttt{train\_loader} is an object that extracts batches with their corresponding targets from a dataset, similar to the PyTorch \texttt{DataLoader}.
In case of classification tasks, the targets $T$ are often represented by a vector of integers, instead of a matrix with the same dimensions as $X$. In such a case a \emph{one-hot encoding} is needed that expands $T$ to a matrix containing zeros and ones. This is taken care of by the \texttt{train\_loader} object.
Note that in our design the neural network parameters $\Theta$ and the optimization functions are stored inside the layers of the MLP $\texttt{M}$. The gradient descent optimizer is
demonstrated in Listing \ref{listing:optimizer-implementation}. It stores a reference to a
parameter \texttt{x} and the corresponding gradient \texttt{Dx}. The update to the parameters is specified in Appendix \ref{appendix:optimization}. A layer can contain multiple parameters.
We support different optimization functions for different parameters. This is achieved using the composite design pattern, as shown in Listing \ref{listing:optimizer-implementation}.
An MLP can be constructed as shown in Listing \ref{listing:mlp-construction}.

\begin{lstfloat}[!ht]
\begin{lstlisting}
def stochastic_gradient_descent(
        M: MultilayerPerceptron,
        epochs: int,
        loss: LossFunction,
        learning_rate: LearningRateScheduler,
        train_loader: DataLoader):

    for epoch in range(epochs):
        lr = learning_rate(epoch)
        for (X, T) in train_loader:
            Y = M.feedforward(X)
            # take the average of the gradients
            N = X.shape[0]  # the batch size
            DY = loss.gradient(Y, T) / N
            M.backpropagate(Y, DY)
            M.optimize(lr)
\end{lstlisting}
\caption{A minimal implementation of stochastic gradient descent in NumPy.}
\label{listing:stochastic-gradient-descent}
\end{lstfloat}

\begin{lstfloat}[!ht]
\begin{lstlisting}
class GradientDescentOptimizer(Optimizer):
    def __init__(self, x, Dx):
        self.x = x
        self.Dx = Dx

    def update(self, eta):
        self.x -= eta * self.Dx

optimizer_W = MomentumOptimizer(layer.W, layer.DW, 0.9)
optimizer_b = NesterovOptimizer(layer.b, layer.Db, 0.9)
L.optimizer = CompositeOptimizer(optimizer_W, optimizer_b)
\end{lstlisting}
\caption{Optimizer Implementation. An optimizer stores references to a parameter \texttt{x} and its gradient \texttt{Dx} with respect to the loss function. Optimizers for multiple parameters are joined using the composite design pattern. The method \texttt{update} performs the actual optimization step.
The parameter \texttt{eta} is the step size or learning rate, and \texttt{L} is an instance of a \texttt{Layer}.
}
\label{listing:optimizer-implementation}
\end{lstfloat}

\begin{lstfloat}[!ht]
\begin{lstlisting}
    M = MultilayerPerceptron()
    input_size = 3072
    output_size = 1024
    act = ReLUActivation()
    layer = ActivationLayer(input_size, output_size, act)
    layer.set_optimizer('Momentum(0.9)')
    layer.set_weights('Xavier')   
    M.layers.append(layer)
    ...
\end{lstlisting}
\caption{Multilayer perceptron construction. The functions \texttt{set\_optimizer} and \texttt{set\_weigths} are offered as convenience functions to simplify the work. It is also possible to
initialize the attributes manually.}
\label{listing:mlp-construction}
\end{lstfloat}

\subsection{The softmax layer} \label{section:softmax-layer}
In this section we show how to design and implement the softmax layer. We use it to demonstrate a non-trivial derivation of a backpropagation equation. For an overview of other layers see Appendix \ref{appendix:layers}. Besides this we show how the backpropagation equations are transformed into code, and how their correctness is validated using SymPy.

A softmax layer is a linear layer that uses the \func{softmax} function (\ref{eq:softmax-single}) as its activation function. It normalizes the output of the linear layer into a probability distribution. The feedforward equations for a single input row vector $x \in \Reals^{1 \times \const{D}}$ are given by
\begin{align}
\begin{cases} \label{eq:softmax-layer-feedforward}
    z &= x W^\top + b \\
    y &= \func{softmax}(z),
\end{cases}
\end{align}
with $y, z \in \Reals^{1 \times \const{K}}$ and with
\begin{align} \label{eq:softmax-single}
    \func{softmax}(z) = \frac{e^z}{e^z \cdot 1_\const{K}}.
\end{align}
Here, $e^z / (e^z \cdot 1_\const{K})$ is in the matrix notation that we use in this paper, see Table \ref{table:matrix-operations}.
It is equivalent to the more common element-wise definition, expressed in the coordinates $z_i$:
\[
\func{softmax}(z)_i = \frac{e^{z_i}}{\sum_{k=1}^\const{K} e^{z_k}} \qquad (1 \leq i \leq \const{K}).
\]
For a mini-batch $X \in \Reals^{\const{N} \times \const{D}}$ the feedforward equations are given by
\begin{align}
\begin{cases}
    Z &= X W^\top + 1_\const{N} \cdot b \\
    Y &= \func{softmax}(Z),
\end{cases}
\end{align}
with $Y, Z \in \Reals^{\const{N} \times \const{K}}$ and
\begin{align}
\label{eq:softmax-multiple}
    \func{softmax}(Z) = e^Z \odot \left( \frac{1}{e^Z \cdot 1_\const{K}}  \cdot 1_\const{K}^\top \right).
\end{align}
As an example we will now derive the equation for the gradient $\Gradient Z$.
Whenever we apply a rule in a derivation, the number of the corresponding equation is put above the
equal sign between parentheses. By the product rule we have
\begin{align} \label{eq:softmax-derivation}
\begin{split}
    &\Partial{z} \func{softmax}(z) =
    \Partial{z} \left( \frac{e^z}{e^z \cdot 1_\const{K}} \right)
    \\
    & = \Partial{z} \left( e^z \cdot \frac{1}{e^z \cdot 1_\const{K}} \right)
    \\
    & \overset{(\ref{eq:product-rule2})}{=}
    \Derivative{e^z}{z} \cdot \frac{1}{e^z \cdot 1_\const{K}} +
    \left( e^z \right)^\top
    \Partial{z} \left( \frac{1}{e^z \cdot 1_\const{K}} \right)
    \\
    &= 
    \func{Diag}(e^z) \cdot \frac{1}{e^z \cdot 1_\const{K}} -
    (e^z)^\top \cdot \frac{e^z}{(e^z \cdot 1_\const{K})^2}
    \\
    &= 
    \func{Diag}\left(\frac{e^z}{e^z \cdot 1_\const{K}}\right) -
    \left(\frac{e^z}{e^z \cdot 1_\const{K}}\right)^\top \cdot
    \frac{e^z}{e^z \cdot 1_\const{K}}
    \\
    &= \func{Diag}(y) -y^\top y,
\end{split}    
\end{align}
with $y = \func{softmax}(z)$.
Now let $y_i, z_i$ be the $i^\text{th}$ row of $Y$, $Z$ for $1 \leq i \leq \const{N}$, hence $y_i = \func{softmax}(z_i)$.
Using the chain rule we find
\[
  \Derivative{\mathcal{L}}{z_i} 
  \overset{(\ref{eq:chain-rule})}{=} 
  \Derivative{\mathcal{L}}{y_i} \Partial{z_i} \func{softmax}(z_i),
\]
hence
\begin{align*}
  \Gradient{z_i} &= \Gradient{y_i} \cdot 
    \left( 
       \func{Diag}(y_i) - y_i^\top y_i
    \right)
  \\  
  &= \Gradient y_i \odot y_i - \Gradient y_i \cdot y_i^\top y_i.
\end{align*}
Using property (\ref{eq:matrix-property5})
we can generalize this into matrix-form:
\begin{align} \label{eq:softmax-layer-backprop1}
  \Gradient Z = (\Gradient Y - \func{diag}(\Gradient Y \cdot Y^\top) \cdot 1_\const{K}^\top) \odot Y.
\end{align}
The remaining backpropagation equations for the softmax layer are
\begin{align} \label{eq:softmax-layer-backprop2}
\begin{split}
\Gradient W &= \Gradient Z^\top \cdot X
\\
\Gradient b &= 1_\const{N}^\top \cdot \Gradient Z
\\
\Gradient X &= \Gradient Z \cdot W.
\end{split}
\end{align}
They are derived in Appendix \ref{appendix:layers}.
Listing \ref{listing:softmax-layer} shows how the feedforward and backpropagation steps are implemented in NumPy, where we have used the matrix operations from Table \ref{table:matrix-operations-implementation} to transform the equations into code.
\begin{lstfloat}[!ht]
\begin{lstlisting}
class SoftmaxLayer(LinearLayer):
  def feedforward(self, X: Matrix) -> Matrix:
    ...
    Z = X @ W.T + row_repeat(b, N)
    Y = softmax(Z)
    ...

  def backpropagate(self, Y: Matrix, DY: Matrix) -> None:
    ...
    DZ = hadamard(Y, DY - column_repeat(diag(DY @ Y.T), K))
    DW = DZ.T @ X
    Db = columns_sum(DZ)
    DX = DZ @ W
    ...
\end{lstlisting}
\caption{Softmax layer implementation in NumPy. We use Table \ref{table:matrix-operations-implementation} to transform equations \ref{eq:softmax-layer-feedforward}, \ref{eq:softmax-layer-backprop1} and \ref{eq:softmax-layer-backprop2} into code.  }
\label{listing:softmax-layer}
\end{lstfloat}
In Listing \ref{listing:gradient-validation} we demonstrate how the backpropagation equations are validated in SymPy, for a small instance $(\const{D} = 3, \const{K} = 2, \const{N} = 2)$ with a simplified squared error loss function. While this does not result in a complete proof, it turns out to be sufficient to detect errors in the equations. Moreover, if an error is detected, it is easy to locate the source of the error by checking the individual steps of the derivation in SymPy. This is a significant improvement over gradient checking using numerical approximations, which does not help to locate the source.
Listing \ref{listing:derivation-validation} shows how the derivation of equation \ref{eq:softmax-derivation} can be validated using SymPy.
\begin{lstfloat}[!ht]
\begin{lstlisting}
    # backpropagation
    DZ = hadamard(Y, DY - column_repeat(diag(DY * Y.T), N))
    DW = DZ.T * X
    Db = columns_sum(DZ)
    DX = DZ * W

    # symbolic differentiation followed by gradient check
    DW1 = gradient(loss(Y), w)
    Db1 = gradient(loss(Y), b)
    DX1 = gradient(loss(Y), x)
    DZ1 = gradient(loss(Y), z)
    self.assertTrue(equal_matrices(DW, DW1))
    self.assertTrue(equal_matrices(Db, Db1))
    self.assertTrue(equal_matrices(DX, DX1))
    self.assertTrue(equal_matrices(DZ, DZ1))
\end{lstlisting}
\caption{Validating gradients in SymPy using symbolic differentiation. The function \texttt{gradient} computes the gradient of a matrix expression (of dimension $1\times 1$ in case of the loss) with respect to a vector of variables, and is implemented using the SymPy symbolic differentiation operator \texttt{diff}. The variables
\texttt{w}, \texttt{b}, \texttt{x}, and \texttt{z} contain the elements of
$W$, $b$, $X$ and $Z$ in equation \ref{eq:softmax-layer-feedforward}.
The function \texttt{equal\_matrices} checks equality of two matrices using the function
\texttt{sympy.simplify}.
}
\label{listing:gradient-validation}
\end{lstfloat}

\begin{lstfloat}[!ht]
\begin{lstlisting}
K = 3
# create a symbolic 1 x K vector
z = Matrix(symarray('z', (1, K), real=True))
y = softmax(z)
e = exp(z)
# rows_sum returns a 1 x 1 Matrix, so we extract the scalar element
f = rows_sum(exp(z))[0, 0]
self.assertTrue(equal_matrices(softmax(z).jacobian(z), Diag(e) / f - (e.T * e) / (f * f)))
self.assertTrue(equal_matrices(softmax(z).jacobian(z), Diag(y) - y.T * y))
\end{lstlisting}
\caption{Checking the derivation in \eqref{eq:softmax-derivation} using SymPy.
The derivation is checked for an arbitrary vector $z$, but with a concrete value for its size $\const{K}$. Note that $e^z \cdot 1_\const{K}$ corresponds to \texttt{rows\_sum(exp(z))} according to Table \ref{table:matrix-operations}.
}
\label{listing:derivation-validation}
\end{lstfloat}

\subsection{Implementation details}
In the implementation of MLPs, one has to take into account that some functions are prone to numerical errors. In this section we list a few of them, and discuss how to solve those issues. Furthermore, we discuss the runtime efficiency of the implementation.

The \func{softmax} function in (\ref{eq:softmax-single}) is prone to underflow and overflow. In practice an equivalent, but numerically more stable version called \func{stable-softmax} is therefore used, see e.g., \citep{10.1093/imanum/draa038}. For single inputs we have
\begin{align}
\label{eq:stable-softmax-single}
\func{stable-softmax}(z) = \frac{e^y}{e^y \cdot 1_\const{K}}
\end{align}
and for multiple outputs
\begin{align}
\label{eq:stable-softmax-multiple}
\func{stable-softmax}(Z) = e^Y \odot \left( \frac{1}{e^Y \cdot 1_\const{K}}  \cdot 1_\const{K}^\top \right),
\end{align}
where
\[
    \begin{cases*}
    y = z - \max(z) \cdot 1_\const{K}^\top
    \\
    Y = Z - \max(Z)_\text{row} \cdot 1_\const{K}^\top.
    \end{cases*}
\]
Note that
\ref{eq:stable-softmax-single}
is the stable version of equation
\ref{eq:softmax-single}
and
\ref{eq:stable-softmax-multiple}
is the stable version of equation
\ref{eq:softmax-multiple}.
Another example is the operation $X^{-1/2}$ that is needed for batch normalization (Appendix~\ref{appendix:batch-normalization}). This operation is unstable for values near zero.
A common solution for this problem is to compute $(X + \varepsilon)^{-1/2}$ instead, meaning that a small positive value $\varepsilon$ is added\footnote{In Keras the value $\varepsilon$ is a parameter of batch normalization. } to the entries of $X$. Finally, the operation $\log(\sigma(X))$ in the logistic cross-entropy loss function
(\ref{eq:logistic-cross-entropy-loss}) needs a special implementation. For this reason the functions \func{inv\_sqrt} and \func{log\_sigmoid} were added to the matrix operations in Table \ref{table:matrix-operations}.

To create a multilayer perceptron implementation based on this paper, consider the following.
Firstly, choose a suitable matrix library, and use it to implement and test the matrix operations in Table \ref{table:matrix-operations}. Secondly, literally implement the layer equations given in Appendix \ref{appendix:layers}. This should result in a functionally correct implementation.
However, for an efficient implementation the matrix library should support hardware acceleration, and preferably also techniques like lazy evaluation and removal of temporaries. The latter are needed to avoid explicit calculation of intermediate values like $1_\const{N} \cdot b$ in the feedforward pass of a linear layer \eqref{eq:linear-layer-feedforward}. Furthermore, profiling of the performance is needed. In particular, more complicated expressions, e.g., in batch normalization \eqref{eq:batch-normalization-backpropagation}, may result in bottlenecks that can be avoided by rewriting them.
The equations and the corresponding code as presented here do give good results for the platforms we tried (see next section), but there may be room for improvement.

\section{Experimental Results} 
\label{section:experimental-results}
\begin{table*}[tbh]
\centering
\caption{Performance comparison of our implementations using the MNIST and CIFAR-10
datasets. The baseline is a native implementation in PyTorch based on an \texttt{nn.Module}. 
In all cases an MLP is used with two hidden layers of sizes 1024 and 512. The results are for one epoch of training with batch size 100 and a constant learning rate of 0.01. The initial weights are the same for each experiment.
The experiments were executed on an an Intel Core i7-6700 system with four CPU cores.
}
\label{table:experiments}
\begin{tabular}{lccccc}
\toprule
Tool & Dataset & Loss & Train Accuracy & Test Accuracy & Time (s) \\
\midrule
PyTorch-Native & MNIST & 0.204 & 0.940 & 0.939 & 3.518 \\
\midrule
Nerva-C++ & MNIST & 0.203 & 0.940 & 0.939 & 3.138 \\
Nerva-Python & MNIST & 0.204 & 0.940 & 0.939 & 3.507 \\
Nerva-PyTorch & MNIST & 0.204 & 0.940 & 0.939 & 4.909 \\
Nerva-TensorFlow & MNIST & 0.204 & 0.940 & 0.939 & 12.818 \\
Nerva-JAX & MNIST & 0.204 & 0.940 & 0.939 & 11.863 \\
Nerva-NumPy & MNIST & 0.204 & 0.940 & 0.939 & 10.630 \\
\midrule
PyTorch-Native & CIFAR-10 & 1.649 & 0.410 & 0.412 & 7.885 \\
\midrule
Nerva-C++ & CIFAR-10 & 1.640 & 0.417 & 0.418 & 7.423 \\
Nerva-Python & CIFAR-10 & 1.650 & 0.413 & 0.413 & 8.073 \\
Nerva-PyTorch & CIFAR-10 & 1.643 & 0.415 & 0.417 & 11.277 \\
Nerva-TensorFlow & CIFAR-10 & 1.657 & 0.407 & 0.407 & 18.941 \\
Nerva-JAX & CIFAR-10 & 1.646 & 0.417 & 0.415 & 19.707 \\
Nerva-NumPy & CIFAR-10 & 1.670 & 0.398 & 0.398 & 20.985 \\
\bottomrule
\end{tabular}
\end{table*}

Our Python implementations are available as PyPI \citep{PyPI} packages
\texttt{nerva-jax},
\texttt{nerva-numpy},
\texttt{nerva-sympy},
\texttt{nerva-tensorflow}, and
\texttt{nerva-torch}, and on GitHub \citep{Wesselink_Nerva_library}.
To validate these implementations, we perform experiments with the MNIST \citep{DBLP:journals/spm/Deng12} and CIFAR-10 \citep{krizhevsky2009learning} datasets. As a baseline we use a native PyTorch model derived from \texttt{nn.Module}, in combination with the \texttt{nn.CrossEntropyLoss} loss function. It is compared with six of our own implementations: our C++ implementation (Nerva-C++) and its Python bindings (Nerva-Python), and our Python implementations in PyTorch, TensorFlow, JAX and NumPy. Note that all our implementations are based on Table \ref{table:matrix-operations} with matrix-form operations.

We use an MLP with two hidden layers of sizes 1024 and 512. The first two layers of the MLP use the ReLU activation function, while the last layer is a linear layer without activation function. Since both experiments are about classification tasks, we use softmax cross-entropy for the loss function. In all cases we have used our own stochastic gradient descent implementation, similar to Listing \ref{listing:stochastic-gradient-descent}.
The initial weights are generated using Xavier, and they are the same for each experiment.
We measure the net training time of one epoch, loading of the dataset is not included.

The results of the experiments can be found in Table \ref{table:experiments}. In both cases our C++ implementation is slightly faster than the native PyTorch implementation. This should not come as a surprise, since the usage of a scripting language like Python usually incurs overhead. However, this demonstrates that an implementation in terms of our table with matrix operations does not necessarily lead to a performance loss.
Our reference implementation in PyTorch is about 40\% slower than a native PyTorch script that uses a \texttt{torch.nn.Module}. This can be seen as the price one has to pay for an explicit implementation of backpropagation in Python. The NumPy, TensorFlow and JAX implementations are significantly slower. The most likely explanation for this is that they are not using a highly optimized library like Intel MKL for matrix computations on the CPU.

\section{Discussion and Conclusions} \label{section:conclusions}
This paper presents an implementation of multilayer perceptrons (MLPs) with a focus on clarity and transparency.
We provide detailed specifications of all the necessary equations in explicit batch matrix-form,
which ease implementation and facilitate analysis.
The mathematical equations are realized through a small, consistent set of matrix primitives,
summarized in tables that bridge formal equations and code.
This uniform approach contributes to more standardized, maintainable, and transparent implementations of neural networks.

Implementations were made in popular Python frameworks, including PyTorch, TensorFlow, and JAX,
and in C++ using Eigen combined with the Intel MKL library.
All these implementations adhere closely to the mathematical specifications,
ensuring correctness and consistency across frameworks.

Our Python implementations fully expose the internal execution of multilayer perceptrons, including all steps of backpropagation. Care has been taken to make the code as simple and readable as possible, in order to provide a valuable resource for students and researchers. A large part of the equations has been derived and validated symbolically using the SymPy library, which accelerates development and ensures mathematical correctness. Experiments with our PyTorch implementation show that it runs approximately 40\% slower than a native PyTorch \texttt{nn.Module} implementation.
For research and educational use, this is a reasonable trade-off for gaining full control and transparency over every computational detail.

We provide an efficient C++ implementation as part of the Nerva library.
Initial experiments indicate that this approach yields competitive performance.
In general, the Eigen library produces efficient code for matrix expressions,
though some expressions required slight reformulation to ensure forwarding
of matrix products to optimized MKL routines. The structured, equation-driven approach advocated in this paper is valuable for understanding and improving computational efficiency, and it naturally extends to other backends, such as GPUs.

A particularly demanding domain for efficient implementation is that of sparse neural networks. Current mainstream frameworks typically emulate sparsity by applying binary masks to dense tensors, which provides no real savings in memory or computation.
In contrast, our C++ backend implements true sparse matrix operations, guided by the same batch matrix-form equations used for dense networks. This makes it possible to analyze correctness and performance at the level of individual matrix operations and to identify bottlenecks in Eigen and MKL. As shown in our previous work \citep{wesselink2024nervatrulysparseimplementation}, this approach enables targeted optimizations and substantial speedups at high sparsity levels. Explicit matrix-form backpropagation thus forms a solid foundation not only for transparent dense
implementations, but also for scaling truly sparse architectures.

In summary, this work contributes:
(1) a complete derivation of batch matrix-form backpropagation for MLPs,
(2) symbolic validation of all gradient equations using SymPy,
(3) uniform implementations across major Python frameworks and a high-performance C++ backend, and (4) demonstration of how explicit matrix formulations enable structured reasoning and efficient sparse computation. Together, these results provide a validated, extensible foundation for both education and research in neural network implementation.

The scope of this paper is limited to standard neural network layers.
Future extensions may include convolutional, pooling, and transformer layers,
continuing the same philosophy of explicit matrix-form specification, symbolic validation, and uniform implementation across frameworks.
The minimal set of matrix primitives developed here provides a basis for such extensions. Further work may also explore GPU implementations and distributed computations, where explicit control over matrix operations can guide new optimizations for both dense and sparse networks.

\bibliography{main}
\clearpage

\appendix

\section{Layer equations} \label{appendix:layers}
A major contribution of this paper is the following overview of the feedforward and backpropagation equations of layers \textbf{in matrix-form}. For each of the equations, the corresponding Python implementation is given. For several equations a derivation is included, while the correctness of the equations and derivations has been validated using SymPy. All layers are implemented in our Nerva Python packages, and a validation of all equations and derivations can be found in the \texttt{tests} directory of the \texttt{nerva-sympy} package. We consider input batches $X$ in row layout, i.e., each row represents a single example. We use the following notations:
\begin{itemize}
    \item $X \in \Reals^{\const{N} \times \const{D}}$ is the input batch, where $\const{N}$ is the number of examples and $\const{D}$ is the input dimension.
    \item $Y \in \Reals^{\const{N} \times \const{K}}$ is the output batch, where $\const{K}$ is the output dimension.
    \item $W \in \Reals^{\const{K} \times \const{D}}$ is the weight matrix, which maps the input features to the output features.
    \item $b \in \Reals^{1 \times \const{K}}$ is the bias vector.
    \item $Z \in \Reals^{\const{N} \times \const{K}}$ is a matrix with intermediate values.
    \item $\beta, \gamma, \Sigma \in \Reals^{1 \times \const{K}}$ are the parameters of batch normalization.
    \item $R \in \Reals^{\const{K} \times \const{D}}$ is a dropout matrix.
\end{itemize}
Each of these matrices has a gradient with the same dimensions, denoted using the same symbol preceded by $\Gradient{}$, e.g., $\Gradient{X}$ is the gradient corresponding to $X$. The implementation uses the same names. The input \texttt{X}, and layer parameters like \texttt{W}, \texttt{b} and their gradients \texttt{DX}, \texttt{DW}, \texttt{Db} are stored in the attributes of a layer. The only exception is the output \texttt{Y} and its gradient \texttt{DY}, which are stored in the \texttt{X} and \texttt{DX} attributes of the next layer. Fig. \ref{fig:layers} explains this in more detail.

\paragraph{linear layer}
\[
\begin{array}{@{} *{2}{L} @{}}
\textsc{feedforward equations} & \textsc{backpropagation equations} \\
\addlinespace[1ex]
\begin{aligned}[t]
Y &= X W^\top + 1_\const{N} \cdot b
\end{aligned}
&
\begin{aligned}[t]
\Gradient W &= \Gradient Y^\top \cdot X
\\
\Gradient b &= 1_\const{N}^\top \cdot \Gradient Y
\\
\Gradient X &= \Gradient Y \cdot W
\end{aligned}
\\
\addlinespace[2ex]
\begin{minipage}[t]{0.5\textwidth}
\small\begin{lstlisting}
Y = X * W.T + row_repeat(b, N)
\end{lstlisting}
\end{minipage}
&
\begin{minipage}[t]{0.5\textwidth}
\small\begin{lstlisting}
DW = DY.T * X
Db = columns_sum(DY)
DX = DY * W
\end{lstlisting}
\end{minipage} \\
\end{array}
\]

\paragraph{activation layer}
Let $\func{act}: \Reals \rightarrow \Reals$ be an activation function, for example \func{relu}.
\[
\begin{array}{@{} *{2}{L} @{}}
\textsc{feedforward equations} & \textsc{backpropagation equations} \\
\addlinespace[1ex]
\begin{aligned}[t]
Z &= X W^\top + 1_\const{N} \cdot b \\
Y &= \func{act}(Z)
\end{aligned}
&
\begin{aligned}[t]
\Gradient Z &= \Gradient Y \odot \func{act}'(Z)
\\
\Gradient W &= \Gradient Z^\top \cdot X
\\
\Gradient b &= 1_\const{N}^\top \cdot \Gradient Z
\\
\Gradient X &= \Gradient Z \cdot W
\end{aligned}
\\
\addlinespace[2ex]
\begin{minipage}[t]{0.5\textwidth}
\small\begin{lstlisting}
Z = X * W.T + row_repeat(b, N)
Y = act(Z)
\end{lstlisting}
\end{minipage}
&
\begin{minipage}[t]{0.5\textwidth}
\small\begin{lstlisting}
DZ = hadamard(DY, act.gradient(Z))
DW = DZ.T * X
Db = columns_sum(DZ)
DX = DZ * W
\end{lstlisting}
\end{minipage} \\
\end{array}
\]

\paragraph{softmax layer}
\[
\begin{array}{@{} *{2}{L} @{}}
\textsc{feedforward equations} & \textsc{backpropagation equations} \\
\addlinespace[1ex]
\begin{aligned}[t]
Z &= X W^\top + 1_\const{N} \cdot b
\\
Y &= \func{softmax}(Z)
\end{aligned}
&
\begin{aligned}[t]
\Gradient Z &= Y \odot (\Gradient Y - \func{diag}(\Gradient Y \cdot Y^\top) \cdot 1_\const{K}^\top)
\\
\Gradient W &= \Gradient Z^\top \cdot X
\\
\Gradient b &= 1_\const{N}^\top \cdot \Gradient Z
\\
\Gradient X &= \Gradient Z \cdot W
\end{aligned}
\\
\addlinespace[2ex]
\begin{minipage}[t]{0.5\textwidth}
\small\begin{lstlisting}
Z = X * W.T + row_repeat(b, N)
Y = softmax(Z)
\end{lstlisting}
\end{minipage}
&
\begin{minipage}[t]{0.5\textwidth}
\small\begin{lstlisting}
DZ = hadamard(Y, DY - column_repeat(diag(DY * Y.T), K))
DW = DZ.T * X
Db = columns_sum(DZ)
DX = DZ * W
\end{lstlisting}
\end{minipage} \\
\end{array}
\]

\paragraph{log-softmax layer}
\[
\begin{array}{@{} *{2}{L} @{}}
\textsc{feedforward equations} & \textsc{backpropagation equations} \\
\addlinespace[1ex]
\begin{aligned}[t]
Z &= X W^\top + 1_\const{N} \cdot b
\\
Y &= \func{logsoftmax}(Z)
\end{aligned}
&
\begin{aligned}[t]
\Gradient Z &= \Gradient Y - \func{softmax}(Z) \odot (\Gradient Y \cdot 1_\const{K} \cdot 1_\const{K}^\top )
\\
\Gradient W &= \Gradient Z^\top \cdot X
\\
\Gradient b &= 1_\const{N}^\top \cdot \Gradient Z
\\
\Gradient X &= \Gradient Z \cdot W
\end{aligned}
\\
\addlinespace[2ex]
\begin{minipage}[t]{0.5\textwidth}
\small\begin{lstlisting}
Z = X * W.T + row_repeat(b, N)
Y = log_softmax(Z)
\end{lstlisting}
\end{minipage}
&
\begin{minipage}[t]{0.5\textwidth}
\small\begin{lstlisting}
DZ = DY - hadamard(softmax(Z), column_repeat(rows_sum(DY), K))
DW = DZ.T * X
Db = columns_sum(DZ)
DX = DZ * W
\end{lstlisting}
\end{minipage} \\
\end{array}
\]

\paragraph{batch normalization layer} \label{appendix:batch-normalization}
\[
\begin{array}{@{} *{2}{L} @{}}
\textsc{feedforward equations} & \textsc{backpropagation equations} \\
\addlinespace[1ex]
\begin{aligned}[t]
R &= X - \frac{1_\const{N} \cdot 1_\const{N}^\top}{\const{N}} \cdot X
\\
\Sigma &= \frac{1}{\const{N}} \cdot \func{diag}(R^\top R)^\top
\\
Z &= (1_\const{N} \cdot \Sigma^{-\frac{1}{2}}) \odot R
\\
Y &= (1_\const{N} \cdot \gamma) \odot Z + 1_\const{N} \cdot \beta
\end{aligned}
&
\begin{aligned}[t]
\Gradient Z &= (1_\const{N} \cdot \gamma) \odot \Gradient Y
\\
\Gradient \beta &= 1_\const{N}^\top \cdot \Gradient Y
\\
\Gradient \gamma &= 1_\const{N}^\top \cdot (Z \odot \Gradient Y)
\\
\Gradient X
     &= \left(\frac{1}{\const{N}} \cdot 1_\const{N} \cdot \Sigma^{-\frac{1}{2}}\right)
\\
     &\odot
       \Big(
          (\const{N} \cdot \mathbb{I}_\const{N}
           - 1_\const{N} \cdot 1_\const{N}^\top) \cdot \Gradient Z
\\[-0.3ex]
     &\qquad
           {}- Z \odot (1_\const{N} \cdot \func{diag}(Z^\top \cdot \Gradient Z)^\top)
       \Big)
\end{aligned}
\\
\addlinespace[2ex]
\begin{minipage}[t]{0.5\textwidth}
\small\begin{lstlisting}
R = X - row_repeat(columns_mean(X), N)
Sigma = diag(R.T * R).T / N
inv_sqrt_Sigma = inv_sqrt(Sigma)
Z = hadamard(row_repeat(inv_sqrt_Sigma, N), R)
Y = hadamard(row_repeat(gamma, N), Z) + row_repeat(beta, N)
\end{lstlisting}
\end{minipage}
&
\begin{minipage}[t]{0.5\textwidth}
\small\begin{lstlisting}
DZ = hadamard(row_repeat(gamma, N), DY)
Dbeta = columns_sum(DY)
Dgamma = columns_sum(hadamard(DY, Z))
DX = hadamard(row_repeat(inv_sqrt_Sigma / N, N), (N * identity(N) - ones(N, N)) * DZ - hadamard(Z, row_repeat(diag(Z.T * DZ).T, N)))
\end{lstlisting}
\end{minipage} \\
\end{array}
\]

\paragraph{linear dropout layer}
\[
\begin{array}{@{} *{2}{L} @{}}
\textsc{feedforward equations} & \textsc{backpropagation equations} \\
\addlinespace[1ex]
\begin{aligned}[t]
Y &= X (W \odot R)^\top + 1_\const{N} \cdot b
\end{aligned}
&
\begin{aligned}[t]
\Gradient W &= (\Gradient Y^\top \cdot X) \odot R
\\
\Gradient b &= 1_\const{N}^\top \cdot \Gradient Y
\\
\Gradient X &= \Gradient Y (W \odot R)
\end{aligned}
\\
\addlinespace[2ex]
\begin{minipage}[t]{0.5\textwidth}
\small\begin{lstlisting}
Y = X * hadamard(W, R).T + row_repeat(b, N)
\end{lstlisting}
\end{minipage}
&
\begin{minipage}[t]{0.5\textwidth}
\small\begin{lstlisting}
DW = hadamard(DY.T * X, R)
Db = columns_sum(DY)
DX = DY * hadamard(W, R)
\end{lstlisting}
\end{minipage} \\
\end{array}
\]

\paragraph{activation dropout layer}
\[
\begin{array}{@{} *{2}{L} @{}}
\textsc{feedforward equations} & \textsc{backpropagation equations} \\
\addlinespace[1ex]
\begin{aligned}[t]
Z &= X (W \odot R)^\top + 1_\const{N} \cdot b
\\
Y &= \func{act}(Z)
\end{aligned}
&
\begin{aligned}[t]
\Gradient Z &= \Gradient Y \odot \func{act}'(Z)
\\
\Gradient W &= (\Gradient Z^\top \cdot X) \odot R
\\
\Gradient b &= 1_\const{N}^\top \cdot \Gradient Z
\\
\Gradient X &= \Gradient Z (W \odot R)
\end{aligned}
\\
\addlinespace[2ex]
\begin{minipage}[t]{0.5\textwidth}
\small\begin{lstlisting}
Z = X * hadamard(W, R).T + row_repeat(b, N)
Y = act(Z)
\end{lstlisting}
\end{minipage}
&
\begin{minipage}[t]{0.5\textwidth}
\small\begin{lstlisting}
DZ = hadamard(DY, act.gradient(Z))
DW = hadamard(DZ.T * X, R)
Db = columns_sum(DZ)
DX = DZ * hadamard(W, R)
\end{lstlisting}
\end{minipage} \\
\end{array}
\]

\subsection{Derivations}
In this section we give some derivations of the backpropagation equations. We give some applications of the product rule and chain rule for vector functions, and show how some properties on the rows and columns of a matrix can be generalized into matrix-form.
\vspace{0.3cm}

\begin{lemma}[Product Rule for vector functions] \label{lemma:product-rule}
The product rule for scalar functions $u$ and $v$ is given by
\[
(u \cdot v)^{\prime} = u^{\prime} \cdot v + u \cdot v^{\prime}.
\]
It can be generalized to vector functions, but the result is sensitive to the orientation of the operands. Below we give four concrete applications of the product rule for vector functions.
Let $x \in \Reals^{p}$, $A \in \Reals^{m \times n}$, and $h(x) = f(x) g(x)$ for $m, n, p \in \mathbb{N}^{+}$.
\begin{table}[h]
\begin{alignat}{2}
&
\text{ Let } f(x) \in \Reals^{n \times 1},
\text{ and let } g(x) \in \Reals, 
\text{ then }
\Derivative{h}{x} = \Derivative{f}{x} g + f \Derivative{g}{x}.
\label{eq:product-rule1}
\\ \addlinespace[1ex]
&
\text{ Let } f(x) \in \Reals^{1 \times n},
\text{ and let } g(x) \in \Reals,
\text{ then }
\Derivative{h}{x} = \Derivative{f}{x} g + f^\top \Derivative{g}{x}.
\label{eq:product-rule2}
\\
&
\text{ Let } f(x) = A,
\text{ and let } g(x) \in \Reals^{n \times 1},
\text{ then }
\Derivative{h}{x} = A \Derivative{g}{x}.
\label{eq:product-rule3}
\\
&
\text{ Let } f(x) \in \Reals^{1 \times m},
\text{ and let } g(x) = A,
\text{ then }
\Derivative{h}{x} = A^\top \Derivative{f}{x}.
\label{eq:product-rule4}
\end{alignat}
\label{table:product-rule}
\end{table}
\end{lemma}

\begin{lemma}[Chain rule for vector functions] \label{lemma:chain-rule}
    Let $f: \Reals^n \rightarrow \Reals^m$, let 
    $g: \Reals^m \rightarrow \Reals^p$,
    and let $h(x) = g(y)$ with $y = f(x)$. Then we have
    \begin{equation} \label{eq:chain-rule}
       \frac{\partial h}{\partial x} = 
       \frac{\partial g}{\partial y} \cdot
       \frac{\partial f}{\partial x}
    \end{equation}
    Note that this equations holds irrespective of whether $f(x)$ is a row or a column vector.
\end{lemma}
\vspace{0.3cm}

\begin{property}[Matrix properties]
Let $X, Y, Z \in \Reals^{m \times n}$. We denote the $i^\text{th}$ row of matrix $A$ as $a_i$ and the $j^\text{th}$ column of matrix $A$ as $a^j$. The element at position $(i,j)$ of $A$ is denoted as $a_{ij}$. Below we give a number of properties on the rows and columns of matrices, and the generalization of these properties into matrix-form.

\begin{alignat}{2}
&\text{If }
  z^j = x^j \cdot (x^j)^\top \cdot y^j \quad (1 \leq j \leq n),
&&\text{ then }
  Z = X \odot (1_m \cdot \func{diag}(X^\top \cdot Y)^\top).
\label{eq:matrix-property1}
\\  
&\text{If }
  z^j = 1_m \cdot (x^j)^\top \cdot y^j \quad (1 \leq j \leq n),
&&\text{ then }
  Z = 1_m \cdot \func{diag}(X^\top Y)^\top.
\label{eq:matrix-property2}
\\  
&\text{If }
  z^j = x^j \cdot 1_m^\top \cdot y^j \quad (1 \leq j \leq n),
&&\text{ then }
  Z = X \odot (1_m \cdot 1_m^\top \cdot Y)
\label{eq:matrix-property3}
\\  
&\text{If }
  z_i = x_i \cdot y_i^\top \cdot y_i \quad (1 \leq i \leq m),
&&\text{ then }
  Z = (\func{diag}(X \cdot Y^\top) \cdot 1_n^\top) \odot Y
\label{eq:matrix-property5}
\\  
&\text{If }
  z_i = x_i \cdot y_i^\top \cdot 1_n^\top \quad (1 \leq i \leq m),
&&\text{ then }
  Z = \func{diag}(X \cdot Y^\top) \cdot 1_n^\top
\label{eq:matrix-property6}
\\  
&\text{If }
  z_i = x_i \cdot 1_n \cdot y_i \quad (1 \leq i \leq m),
&&\text{ then }
  Z = (X \cdot 1_n \cdot 1_n^\top) \odot Y
\label{eq:matrix-property7}
\end{alignat}
\end{property}

\noindent
All properties have been validated with SymPy. Naturally there is a lot of symmetry between the row and column properties. By means of example we will prove equation \eqref{eq:matrix-property5}. From $z_i = x_i \cdot y_i^\top \cdot y_i$ we derive that $z_{ij} = (x_i \cdot y_i^\top) y_{ij}$ for $1 \leq j \leq n$. Hence we can write $Z = R \odot Y$, where $R$ is defined using $r_{ij} = (x_i \cdot y_i^\top)$. We observe that $\func{diag}(X \cdot Y^\top) = (x_1 \cdot y_1^\top, \ldots, x_m \cdot y_m^\top)^\top$. From the definition of $R$ we can see that it consists of $n$ copies of the column vector $\func{diag}(X \cdot Y^\top)$. Hence we have $R = \func{diag}(X \cdot Y^\top) \cdot 1_n^\top$.

\paragraph{linear layer}
We have $Y = X \cdot W^\top + 1_\const{N} \cdot b$. Let $x_i, y_i, b_i$ be the $i^\text{th}$ row of $X$, $Y$, $1_\const{N} \cdot b$ for $1 \leq i \leq \const{N}$, hence $y_i = x_i W^\top + b_i$, where $b_i = b$. Furthermore, let $w_j$ be the $j^\text{th}$ row of $W$, and let $e^i$ be the $i^\text{th}$ column of the unit matrix $\mathbb{I}_\const{K}$. Let $\mathcal{L}(Y) = \sum_{i=1}^\const{N} \mathcal{L}(y_i)$ be the corresponding loss. We calculate
\begin{align}
  \label{eq:dLdx}
  \Derivative{\mathcal{L}}{x_i} 
  &\overset{(\ref{eq:chain-rule})}{=} \sum_{n=1}^\const{N} \Derivative{\mathcal{L}}{y_n} \Derivative{y_n}{x_i}
  = \Derivative{\mathcal{L}}{y_i} \Derivative{y_i}{x_i}
  \overset{(\ref{eq:product-rule4})}{=} \Derivative{\mathcal{L}}{y_i} \cdot W,
  \text{ hence } \Gradient{x_i} = \Gradient{y_i} \cdot W
  \\
  \label{eq:dLdb}
  \Derivative{\mathcal{L}}{b} 
  &\overset{(\ref{eq:chain-rule})}{=} \sum_{n=1}^\const{N} \Derivative{\mathcal{L}}{y_n} \Derivative{y_n}{b}
  \overset{(\ref{eq:product-rule4})}{=} \sum_{n=1}^\const{N} \Derivative{\mathcal{L}}{y_n} \cdot \mathbb{I}_K
  = \sum_{n=1}^\const{N} \Derivative{\mathcal{L}}{y_n}, 
  \text{ hence } \Gradient{b} = 1_\const{N}^\top \cdot \Gradient{Y}
\end{align}
For the gradient of $W$ we calculate the derivative with respect to an arbitrary entry \(w_{i j}\).
From the definition of $Y$ we derive
\[
\frac{\partial y_{n k}}{\partial w_{i j}}
=
\frac{\partial (\sum_{d=1}^{D} x_{n d}\,w_{k d}+b_k)}{\partial w_{i j}}
=
\begin{cases}
x_{n j}, & \text{if } k=i,\\[4pt]
0,       & \text{if } k\neq i.
\end{cases}
\]
Using this, we calculate
\begin{equation}
  \label{eq:dLdw}
\frac{\partial\mathcal{L}}{\partial w_{i j}}
\;=\;
\sum_{n=1}^{N}\sum_{k=1}^{K}
\frac{\partial\mathcal{L}}{\partial y_{n k}}\,
\frac{\partial y_{n k}}{\partial w_{i j}}
= \sum_{n=1}^{N} \frac{\partial\mathcal{L}}{\partial y_{n i}} \; x_{n j}.
\end{equation}
The equations \ref{eq:dLdx}, \ref{eq:dLdb} and \ref{eq:dLdw} can be generalized to matrix equations: $\Gradient{X} = \Gradient{Y} \cdot W$, $\Gradient{b} = 1_\const{N}^\top \cdot \Gradient{Y}$, and $\Gradient{W} = \Gradient Y^\top \cdot X$.

\paragraph{log-softmax layer}
We have $Y = \func{log-softmax}(Z)$.
Let $y_i, z_i$ be the $i^\text{th}$ row of $Y$, $Z$ for $1 \leq i \leq \const{N}$, hence $y_i = \func{log-softmax}(z_i)$.
We calculate
\begin{align*}
  \Derivative{\mathcal{L}}{z_i} 
  & \overset{(\ref{eq:chain-rule})}{=} 
    \Derivative{\mathcal{L}}{y_i} \Partial{z_i} \func{log-softmax}(z_i)
  \overset{(\ref{eq:log-softmax-derivation})}{=}
  \Derivative{\mathcal{L}}{y_i}
    \left( 
       \mathbb{I}_K - 1_\const{K} \cdot \func{softmax}(z_i)
    \right)
  \\
  & \Longleftrightarrow~~ \Gradient{z_i} = \Gradient{y_i} \cdot 
      \left( 
       \mathbb{I}_K - 1_\const{K} \cdot \func{softmax}(z_i)
    \right)
  = \Gradient y_i - \Gradient y_i \cdot 1_\const{K} \cdot \func{softmax}(z_i).  
\end{align*}
This can be generalized to matrices using property \ref{eq:matrix-property7}:
\begin{align}
\Gradient Z &= \Gradient Y - \func{softmax}(Z) \odot
(\Gradient Y \cdot 1_\const{K} \cdot 1_\const{K}^\top)
\end{align}

\paragraph{batch normalization layer}
We will derive the equation for $\Gradient X$, which is the most complicated one.
Following the approach of \citep{yeh-batch-norm}, we first derive the equations for a single column. Let $x^j, r^j, z^j \in \Reals^{\const{N} \times 1}$ be the $j^\text{th}$ column of $X$, $R$ and $Z$, and let $\sigma \in \Reals$ be the $j^\text{th}$ element of $\Sigma$, with $1 \leq j \leq \const{D}$. Then we obtain the following equations:
\begin{align}
r^j &= x^j - \frac{1_\const{N} \cdot 1_\const{N}^\top}{\const{N}} \cdot x^j = (\mathbb{I}_{\const{N}} - \frac{1_\const{N} \cdot 1_\const{N}^\top}{\const{N}}) \cdot x^j
\\
\sigma &= \frac{(r^j)^\top r^j}{\const{N}} 
\\
z^j &= r^j \cdot \sigma^{-\frac{1}{2}}
\end{align}
We calculate
\begin{align}
  \Derivative{z^j}{r^j}
  & \overset{(\ref{eq:product-rule1})}{=}
  \Derivative{r^j}{r^j} \cdot \sigma^{-\frac{1}{2}}
  + r^j \cdot \Derivative{\sigma^{-\frac{1}{2}}}{r^j}
  \overset{(\ref{eq:chain-rule})}{=} 
  \sigma^{-\frac{1}{2}} - r^j \cdot 
  \frac{\sigma^{-\frac{3}{2}}}{2} \cdot \Derivative{\sigma}{r^j}
  =
  \sigma^{-\frac{1}{2}} - r^j \cdot 
  \frac{\sigma^{-\frac{3}{2}}}{2} \cdot \left( \frac{2 (r^j)^\top}{\const{N}} \right)
  \\
  &= 
  \frac{\sigma^{-\frac{1}{2}}}{\const{N}}
  \left(
    \const{N} \cdot \mathbb{I}_\const{N} - \sigma^{-1} r^j (r^j)^\top
  \right)
  =
  \frac{\sigma^{-\frac{1}{2}}}{\const{N}}
  \cdot 
  \left( 
    \const{N} \cdot \mathbb{I}_{\const{N}} - z^j (z^j)^\top  
  \right).
\end{align}
Using the chain rule we find
\begin{align}
  \frac{\partial \mathcal{L}}{\partial r^j} 
  & \overset{(\ref{eq:chain-rule})}{=}
  \frac{\partial \mathcal{L}}{\partial z^j}
  \frac{\partial z^j}{\partial r^j},
  \text{ hence }
  \Gradient{r^j} = 
  \frac{\sigma^{-\frac{1}{2}}}{\const{N}}
  \left( 
    \const{N} \cdot \mathbb{I}_{\const{N}} - z^j (z^j)^\top  
  \right)
  \cdot
  \Gradient{z^j}
  \\
  \frac{\partial \mathcal{L}}{\partial x^j} 
  & \overset{(\ref{eq:chain-rule})}{=}
  \frac{\partial \mathcal{L}}{\partial r^j}
  \frac{\partial r^j}{\partial x^j},
  \text{ hence }
  \Gradient x^j 
  = 
  (\mathbb{I}_{\const{N}} - \frac{1_\const{N} \cdot 1_\const{N}^\top}{\const{N}})
  \cdot
  \Gradient r^j,
\end{align}
where one needs to take into account that for a column vector $x$ we have $\Gradient x = (\Derivative{\mathcal{L}}{x})^\top$.
Hence
\begin{align*}
    \Gradient x^j 
    &=
    \frac{\sigma^{-\frac{1}{2}}}{\const{N}}
    \cdot
    (\mathbb{I}_{\const{N}} - \frac{1_\const{N} \cdot 1_\const{N}^\top}{\const{N}})
    \cdot 
    \left( 
       \const{N} \cdot \mathbb{I}_{\const{N}} - z^j \cdot (z^j)^\top 
    \right)
    \cdot
    \Gradient z^j
    \\
    &=
    \frac{\sigma^{-\frac{1}{2}}}{\const{N}}
    \cdot
    (
      \const{N} \cdot \mathbb{I}_{\const{N}}
      - z^j \cdot (z^j)^\top
      - 1_\const{N} \cdot 1_\const{N}^\top
      + \frac{1_\const{N} \cdot 1_\const{N}^\top \cdot z^j \cdot (z^j)^\top}{\const{N}}
    )  
    \cdot
    \Gradient z^j
    \\     
      & \ \ \ \  \{ \; \text{In batch normalization the column sum $1_\const{N}^\top \cdot z^j$ evaluates to zero. } \}
    \\ 
    &=
    \frac{\sigma^{-\frac{1}{2}}}{\const{N}}
    \cdot
    (
      (\const{N} \cdot \mathbb{I}_{\const{N}} - 1_\const{N} \cdot 1_\const{N}^\top) \cdot \Gradient z^j
      - z^j \cdot (z^j)^\top \cdot \Gradient z^j
    )  
\end{align*}
Using property \eqref{eq:matrix-property1} we generalize this into matrix-form:
\begin{align} \label{eq:batch-normalization-backpropagation}
\Gradient X
     &= (\frac{1}{\const{N}} \cdot 1_\const{N} \cdot \Sigma^{-\frac{1}{2}}) ~\odot 
      \left(
              (\const{N} \cdot \mathbb{I}_\const{N} - 1_\const{N} \cdot 1_\const{N}^\top) \cdot \Gradient Z
              -
              Z \odot (1_\const{N} \cdot \func{diag}(Z^\top \cdot \Gradient Z)^\top)
       \right).
\end{align}
For the remaining equations, let 
$y_i, z_i, \beta_i, \gamma_i$ be the $i^\text{th}$ row of $Y$, $Z$, $1_\const{N} \cdot \beta$, and $1_\const{N} \cdot \gamma$ for $1 \leq i \leq \const{N}$, where $\beta_i = \beta$, and $\gamma_i = \gamma$. Hence $y_i = \gamma_i \odot z_i + \beta_i$. From this it follows
\begin{align*}
    & \Derivative{\mathcal{L}}{\beta_i} = \Derivative{\mathcal{L}}{y_i}
    \\
    & \Derivative{\mathcal{L}}{\gamma_i} = \Derivative{\mathcal{L}}{y_i} \odot z_i,
\end{align*}
which we can generalize into matrix-form:
$\Gradient \beta = 1_\const{N}^\top \cdot \Gradient Y$ and
$\Gradient \gamma = 1_\const{N}^\top \cdot (\Gradient Y \odot Z)$.

\section{Implementation of matrix operations} \label{appendix:matrix_operations}
In Table \ref{table:matrix-operations-implementation} we provide implementations in several frameworks of the fundamental matrix-form operations for multilayer perceptrons. See Table~\ref{table:matrix-operations} in the main body of the paper for the corresponding equations and definitions.

\begin{table*}[h]
\scriptsize
\centering
\begin{tabular}{ p{2.2cm}|p{2.6cm}p{2.6cm}p{2.6cm}p{3.6cm} }
 \toprule
 Operation & NumPy + JAX & PyTorch & TensorFlow & Eigen\\
 \midrule
\textsf{zeros(m,n)}            & \textsf{zeros((m,n))}           & \textsf{zeros(m,n)}              & \textsf{zeros([m,n])}                & \textsf{Zero(m,n)}                      \\
\textsf{ones(m,n)}             & \textsf{ones((m,n))}            & \textsf{ones(m,n)}               & \textsf{ones([m,n])}                 & \textsf{Ones(m,n)}                      \\
\textsf{identity(n)}           & \textsf{eye(n)}                 & \textsf{eye(n)}                  & \textsf{eye(n)}                      & \textsf{Identity(n,n)}                  \\
\textsf{X.T}                   & \textsf{X.T}                    & \textsf{X.T}                     & \textsf{X.T}                         & \textsf{X.transpose()}                   \\
\textsf{c * X}                 & \textsf{c * X}                  & \textsf{c * X}                   & \textsf{c * X}                       & \textsf{c * X}                           \\
\textsf{X + Y}                 & \textsf{X + Y}                  & \textsf{X + Y}                   & \textsf{X + Y}                       & \textsf{X + Y}                           \\
\textsf{X - Y}                 & \textsf{X - Y}                  & \textsf{X - Y}                   & \textsf{X - Y}                       & \textsf{X - Y}                           \\
\textsf{X * Z}                 & \textsf{X @ Z}                  & \textsf{X @ Z}                   & \textsf{X @ Z}                       & \textsf{X * Z}                           \\
\textsf{hadamard(X,Y)}         & \textsf{X * Y}                  & \textsf{X * Y}                   & \textsf{multiply(X,Y)}               & \textsf{X.array() * Y.array()}           \\
\textsf{diag(X)}               & \textsf{diag(X)}                & \textsf{diag(X)}                 & \textsf{diag\_part(X)}               & \textsf{X.diagonal()}                    \\
\textsf{Diag(x)}               & \textsf{diag(x)}                & \textsf{diag(x.flatten())}       & \textsf{diag(reshape(x,[-1]))}       & \textsf{x.asDiagonal()}                  \\
\textsf{elements\_sum(X)}      & \textsf{sum(X)}                 & \textsf{sum(X)}                  & \textsf{reduce\_sum(X)}              & \textsf{X.sum()}                         \\
\textsf{column\_repeat(x,n)}   & \textsf{tile(x,(1,n))}          & \textsf{x.repeat(1,n)}           & \textsf{tile(x,[1,n])}               & \textsf{x.replicate(1,n)}               \\
\textsf{row\_repeat(x,m)}      & \textsf{tile(x,(m,1))}          & \textsf{x.repeat(m,1)}           & \textsf{tile(x,[m,1])}               & \textsf{x.replicate(m,1)}               \\
\textsf{columns\_sum(X)}       & \textsf{sum(X,axis=0)}          & \textsf{sum(X,dim=0)}            & \textsf{reduce\_sum(X,axis=0)}       & \textsf{X.colwise().sum()}               \\
\textsf{rows\_sum(X)}          & \textsf{sum(X,axis=1)}          & \textsf{sum(X,dim=1)}            & \textsf{reduce\_sum(X,axis=1)}       & \textsf{X.rowwise().sum()}               \\
\textsf{columns\_max(X)}       & \textsf{max(X,axis=0)}          & \textsf{max(X,dim=0).values}     & \textsf{reduce\_max(X,axis=0)}       & \textsf{X.colwise().maxCoeff()}          \\
\textsf{rows\_max(X)}          & \textsf{max(X,axis=1)}          & \textsf{max(X,dim=1).values}     & \textsf{reduce\_max(X,axis=1)}       & \textsf{X.rowwise().maxCoeff()}          \\
\textsf{columns\_mean(X)}      & \textsf{mean(X,axis=0)}         & \textsf{mean(X,dim=0)}           & \textsf{reduce\_mean(X,axis=0)}      & \textsf{X.colwise().mean()}              \\
\textsf{rows\_mean(X)}         & \textsf{mean(X,axis=1)}         & \textsf{mean(X,dim=1)}           & \textsf{reduce\_mean(X,axis=1)}      & \textsf{X.rowwise().mean()}              \\
\textsf{apply(f,X)}            & \textsf{f(X)}                   & \textsf{f(X)}                    & \textsf{f(X)}                        & \textsf{X.unaryExpr(f)}                  \\
\textsf{exp(X)}                & \textsf{exp(X)}                 & \textsf{exp(X)}                  & \textsf{exp(X)}                      & \textsf{X.array().exp()}                 \\
\textsf{log(X)}                & \textsf{log(X)}                 & \textsf{log(X)}                  & \textsf{log(X)}                      & \textsf{X.array().log()}                 \\
\textsf{reciprocal(X)}         & \textsf{1 / X}                  & \textsf{1 / X}                   & \textsf{reciprocal(X)}               & \textsf{X.array().inverse()}             \\
\textsf{sqrt(X)}               & \textsf{sqrt(X)}                & \textsf{sqrt(X)}                 & \textsf{sqrt(X)}                     & \textsf{X.array().sqrt()}                \\
\textsf{inv\_sqrt(X)} & \textsf{reciprocal(sqrt(X+e))}  & \textsf{reciprocal(sqrt(X+e))}   & \textsf{reciprocal(sqrt(X+e))}       & \textsf{reciprocal(sqrt(X.array()+e))}   \\
\textsf{log\_sigmoid(X)}       & \textsf{-logaddexp(0,-X)}       & \textsf{-softplus(-X)}           & \textsf{-softplus(-X)}               & \textsf{-log1p(exp(-X.array()))}         \\
 \bottomrule
\end{tabular}
\vspace{.8em}
\caption{Implementation of matrix operations in NumPy, JAX, PyTorch, TensorFlow and Eigen.
We assume that \texttt{e} is a given small positive constant that is used to avoid division by zero.
Note that some of the operations are located in Python submodules.
}
\label{table:matrix-operations-implementation}
\end{table*}

\section{Activation functions} \label{appendix:activation-functions}
In this section we give an overview of some commonly used univariate activation functions. These activation functions are typically applied element-wise to the output matrix $Y$ of a neural network.
\[
\begin{array}{L @{\hspace{10pt}} L @{\hspace{20pt}} L}
  \textsc{name} & \textsc{function} & \textsc{derivative}
  \\

  \begin{minipage}[t]{0.3\textwidth}
    \text{ReLU} \\ \text{\citep{Fuk75}} 
  \end{minipage}
  &
  \func{relu}(x) = \max(0, x)
  &
  \func{relu}'(x) = 
 \begin{cases*}
    0 & if $x < 0$ \\
    1 & otherwise
  \end{cases*}
  \\
  \addlinespace[1ex]

  \begin{minipage}[t]{0.3\textwidth}
    \text{Leaky ReLU} \\ \text{\citep{Maas2013RectifierNI}}
  \end{minipage}
  &
  \func{leaky-relu}(x) = \max(\alpha x, x)
  &
\func{leaky-relu}'(x) = 
 \begin{cases*}
    \alpha & if $x < \alpha x$ \\
    1 & otherwise
  \end{cases*}
  \\
  \addlinespace[1ex]

  \text{Hyperbolic tangent}
  &
  \func{tanh}(x) = \frac{e^x - e^{-x}}{e^x + e^{-x}}
  &
  \func{tanh}'(x) = 1 - \func{tanh}(x)^2
  \\
  \addlinespace[1ex]

  \begin{minipage}[t]{0.3\textwidth}
    \text{Sigmoid / logistic function} \\ \text{\citep{hinton12}}
  \end{minipage}
  &
  \sigma(x) = \frac{1}{1 + e^{-x}}
  &
  \sigma'(x) = \sigma(x) (1 - \sigma(x))
\end{array}
\]
\vspace{0.5em}

\noindent
Note that leaky ReLU has a parameter $\alpha \in \mathbb{R}$, usually $0 < \alpha < 1$.

\subsection{Softmax functions}
In this section we give an overview of softmax functions and their derivatives, both for single inputs and for batches. These equations are present in the activation function of softmax layers and in some of the loss functions in Appendix \ref{appendix:loss-functions}. We use the same notation as before, where $x \in \Reals^{1 \times \const{D}}$ is a single input with dimension $\const{D}$, and $X \in \Reals^{\const{N} \times \const{D}}$ is an input batch, where $\const{N}$ is the number of samples in the batch.
We denote the rows of $X$ as $x_1, \ldots, x_\const{N}$.\\

\noindent
Below an overview of softmax functions and their derivatives is given. Equations in matrix-form are only given if they are needed later on. Note that the function $\func{log-softmax}$ is defined using
\[
\func{log-softmax}(x) = \log(\func{softmax}(x)).
\]
\[
\begin{array}{@{} *{2}{L} @{}}
  \textsc{vector functions} & \textsc{matrix functions}
  \\
  
  \addlinespace[1ex]
  \func{softmax}(x) = \frac{e^x}{e^x \cdot 1_\const{D}}
  &
  \func{softmax}(X) = e^X \odot \left( \frac{1}{e^X \cdot 1_\const{D}}  \cdot 1_\const{D}^\top \right)
  \\
  \addlinespace[1ex]

  \func{stable-softmax}(x) = \frac{e^z}{e^z \cdot 1_\const{D}}
  &
  \func{stable-softmax}(X) = e^Z \odot \left( \frac{1}{e^Z \cdot 1_\const{D}}  \cdot 1_\const{D}^\top \right)
  \\
  \addlinespace[1ex]
  
  \func{log-softmax}(x) = x - \log(e^x \cdot 1_\const{D}) \cdot 1_\const{D}^\top
  &
  \func{log-softmax}(X) = X - \log(e^X \cdot 1_\const{D}) \cdot 1_\const{D}^\top
  \\
  \addlinespace[2ex]

  \func{stable-log-softmax}(x) = z - \log(e^z \cdot 1_\const{D}) \cdot 1_\const{D}^\top
  &
  \func{stable-log-softmax}(X) = Z - \log(e^Z \cdot 1_\const{D}) \cdot 1_\const{D}^\top,
\end{array}
\]

\ \\
where

\[
    \begin{cases*}
    z = x - \max(x) \cdot 1_\const{D}^\top 
    \\
    Z = X - \max(X)_\text{row} \cdot 1_\const{D}^\top.
    \end{cases*}
\]

\ \\
\noindent
The derivatives of the softmax functions are given by
\begin{align}
  \frac{\partial}{\partial x} \func{softmax}(x) &= 
  \frac{\partial}{\partial x} \func{stable-softmax}(x) =
  \func{Diag}(\func{softmax}(x)) - \func{softmax}(x)^\top \cdot \func{softmax}(x),
  \\
  \frac{\partial}{\partial x} \func{log-softmax}(x) &= 
  \frac{\partial}{\partial x} \func{stable-log-softmax}(x) =
  \mathbb{I}_D - 1_\const{D} \cdot \func{softmax}(x)
\end{align}

\subsection{Derivations}

\paragraph{log-softmax}
Let $y(x) = \func{softmax}(x)$, then we have
\begin{align} \label{eq:log-softmax-derivation}
\begin{split}
    \Partial{x} \log \left( \func{softmax}(x) \right)
    & \overset{(\ref{eq:chain-rule})}{=}
    \Partial{y} \log(y) \Partial{x} \func{softmax}(x)
    \\
    & \overset{(\ref{eq:softmax-derivation})}{=}
    \func{Diag}\left(\frac{1}{y}\right) \cdot 
        \left( \func{Diag}(y) -y^\top y \right)
    \\
    &= \mathbb{I}_D - \func{Diag}\left(\frac{1}{y}\right) y^\top y
    \\
    &= \mathbb{I}_D - 1_\const{D} \cdot y
    \\
    &= \mathbb{I}_D - 1_\const{D} \cdot \func{softmax}(x).
\end{split}    
\end{align}

\section{Loss functions} \label{appendix:loss-functions}
In this section we give an overview of some loss functions and their gradients, both for single inputs and for batches. We use the following notations:
\begin{enumerate}
    \item $y \in \Reals^{1 \times \const{K}}$ is a single output, where $\const{K}$ is the output dimension.
    \item $t \in \Reals^{1 \times \const{K}}$ is a target for a single output $y$.
    \item $Z \in \Reals^{\const{N} \times \const{K}}$ is an output batch, where $\const{N}$ is the number of examples in the batch.
    \item $T \in \Reals^{\const{N} \times \const{K}}$ contains the targets for an output batch $Y$.
\end{enumerate}

\paragraph{squared error loss}
\begin{align}
\begin{split}
  \mathcal{L}_\text{SE}(y, t) &= (y - t) (y - t)^\top
  \\
  \nabla_y \mathcal{L}_\text{SE}(y, t) &= 2(y - t)
  \\
  \mathcal{L}_\text{SE}(Y, T) &= 1_\const{N}^\top \cdot ((Y - T) \odot (Y - T)) \cdot 1_\const{K}
  \\
  \nabla_Y \mathcal{L}_\text{SE}(Y, T) &= 2(Y - T)
\end{split}  
\end{align}
The mean squared error loss is a scaled version of the squared error loss.
\begin{align*}
  \mathcal{L}_\text{MSE}(Y, T) &= \frac{\mathcal{L}_\text{SE}(Y, T)} {\const{K} \cdot \const{N}}
\end{align*}

\paragraph{cross-entropy loss}
\begin{align}
\begin{split}
  \mathcal{L}_\text{CE}(y, t) &= - t \cdot \log(y)^\top
  \\
  \nabla_y \mathcal{L}_\text{CE}(y, t) &= - t \odot \frac{1}{y}
  \\
  \mathcal{L}_\text{CE}(Y, T) &= - 1_\const{N}^\top \cdot (T \odot \log(Y)) \cdot 1_\const{K}
  \\
  \nabla_Y \mathcal{L}_\text{CE}(Y, T) &= -T \odot \frac{1}{Y}
\end{split} 
\end{align}

\paragraph{softmax cross-entropy loss}
\begin{align}
\begin{split}
  \mathcal{L}_\text{SCE}(y, t) &= - t \cdot \func{log-softmax}(y)^\top
  \\
  \nabla_y \mathcal{L}_\text{SCE}(y, t) &= t \cdot 1_\const{K} \cdot \func{softmax}(y) - t
  \\
  \mathcal{L}_\text{SCE}(Y, T) &= - 1_\const{N}^\top \cdot (T \odot \func{log-softmax}(Y)) \cdot 1_\const{K}
  \\
  \nabla_Y \mathcal{L}_\text{SCE}(Y, T) &= \func{softmax}(Y) \odot (T \cdot 1_\const{K} \cdot 1_\const{K}^\top) - T
\end{split}
\end{align}
Note that if $t$ is a one-hot encoded target (e.g., in case of a classification problem),
it is a vector consisting of one value 1 and all others values 0. In other words
we have $t \cdot 1_\const{K} = 1$, hence in this case the gradients can be simplified to
\begin{align*}
  \nabla_y \mathcal{L}_\text{SCE}(y, t) &= \func{softmax}(y) - t \\
  \nabla_Y \mathcal{L}_\text{SCE}(Y, T) &= \func{softmax}(Y) - T
\end{align*}

\paragraph{logistic cross-entropy loss}
\begin{align}
  \begin{split}
  \label{eq:logistic-cross-entropy-loss}
  \mathcal{L}_\text{LCE}(y, t) &= - t \cdot \log(\sigma(y))^\top
  \\
  \nabla_y \mathcal{L}_\text{LCE}(y, t) &= t \odot \sigma(y) - t
  \\
  \mathcal{L}_\text{LCE}(Y, T) &= - 1_\const{N}^\top \cdot (T \odot \log(\sigma(Y))) \cdot 1_\const{K}
  \\
  \nabla_Y \mathcal{L}_\text{LCE}(Y, T) &= T \odot \sigma(Y) - T
  \end{split}
\end{align}

\paragraph{negative log-likelihood loss}
\begin{align}
\begin{split}
  \mathcal{L}_\text{NLL}(y, t) &= - \log(y \cdot t^\top)
  \\
  \nabla_y \mathcal{L}_\text{NLL}(y, t) &= - \frac{1}{y \cdot t^\top} \cdot t
  \\
  \mathcal{L}_\text{NLL}(Y, T) &= - 1_\const{N}^\top \cdot (\log((Y \odot T) \cdot 1_\const{K}))
  \\
  \nabla_Y \mathcal{L}_\text{NLL}(Y, T) &= -\left( \frac{1}{(Y \odot T) \cdot 1_\const{K}} \cdot 1_\const{K}^\top \right) \odot T
\end{split}
\end{align}

\subsection{Derivations}

\paragraph{cross-entropy loss}
\begin{align}
\begin{split}
\nabla_y \mathcal{L}_\text{CE}(y, t)
&= - \Partial{y} (t \cdot \log(y)^\top)
\overset{(\ref{eq:product-rule3})}{=} - t \cdot \func{Diag}\left(\frac{1}{y}\right)
= - t \odot \frac{1}{y} \\
\end{split}
\end{align}

\paragraph{softmax cross-entropy loss}
\begin{align}
\label{eq:softmax-cross-entropy-derivation}
\begin{split}
\nabla_y \mathcal{L}_\text{SCE}(y, t) 
&= - \Partial{y} (t \cdot \func{log-softmax}(y)^\top)
\overset{(\ref{eq:product-rule3})}{=} - t \cdot \Partial{y} \func{log-softmax}(y) 
\\
& \overset{(\ref{eq:log-softmax-derivation})}{=} -t \cdot \left(\mathbb{I}_K - 1_\const{K} \cdot \func{softmax}(y) \right)
= t \cdot 1_\const{K} \cdot \func{softmax}(y) - t
\end{split}
\end{align}
If the target $t$ is a one-hot encoded vector, we have $t \cdot 1_\const{K} = 1$. In that case the gradient simplifies to
\[
  \nabla_y \mathcal{L}_\text{SCE-one-hot}(y, t) = \func{softmax(y)} - t.
\]
Using property \eqref{eq:matrix-property7} we can generalize equation \eqref{eq:softmax-cross-entropy-derivation} to
\begin{align*}
\nabla_Y \mathcal{L}_\text{SCE}(Y, T)
&= \func{softmax}(Y) \odot (T \cdot 1_\const{K} \cdot 1_\const{K}^\top) - T.
\end{align*}

\paragraph{logistic cross-entropy loss}
\begin{align*}
\nabla_y \mathcal{L}_\text{LCE}(y, t) 
&= \Partial{y}(-t \cdot \log(\sigma(y))^\top)
\overset{(\ref{eq:product-rule3})}{=} -t \cdot \Partial{y}\log(\sigma(y))
\\
&= - t \cdot \func{Diag}(1_\const{K}^\top - \sigma(y))
= t \odot \sigma(y) - t,
\end{align*}
where we use the fact that in the univariate case:
\begin{equation*}
    \dfrac{d}{dt}\log(\sigma(t)) = \dfrac{\sigma'(t)}{\sigma(t)} = \dfrac{\sigma(t) \cdot (1 - \sigma(t))}{\sigma(t)} = 1-\sigma(t).
\end{equation*}

\section{Weight initialization} \label{appendix:weight-initialization}
The initial values of the weights in linear layers need to be chosen carefully, since they may have a large impact on the performance of a neural network \citep{DBLP:journals/air/NarkhedeBS22}.
Typically, these values are randomly generated based on specific probability distributions. In this section we give a few commonly used distributions.

\[
\begin{array}{@{} *{3}{L} @{}}
  \textsc{name} & \textsc{distribution}
  \\
  \addlinespace[1ex]

  \text{Uniform} & U(a,b)
  \\
  \addlinespace[1ex]

  \text{Xavier \citep{pmlr-v9-glorot10a}} & U(- \frac{1}{\sqrt{\const{D}}}, \frac{1}{\sqrt{\const{D}}})
  \\
  \addlinespace[1ex]

  \text{Normalized Xavier \citep{pmlr-v9-glorot10a}} & U(- \frac{\sqrt{6}}{\sqrt{\const{D}+\const{K}}}, \frac{\sqrt{6}}{\sqrt{\const{D}+\const{K}}}),
  \\
  \addlinespace[1ex]

  \text{He \citep{he2015delving}} & \mathcal{N}(0, \sqrt{\frac{2}{\const{D}}}),

\end{array}
\]
where $\const{D}$ is the number of inputs and $\const{K}$ the number of outputs of the layer to which the weight matrix belongs. Furthermore $U(a,b)$ is the uniform distribution on a given interval $[a, b]$, and $\mathcal{N}(\mu,\sigma)$ is the normal distribution with mean $\mu$ and standard deviation $\sigma$.

Unlike weights, the initial values of bias vectors are typically set to a small constant or zero rather than drawn from probability distributions. However, the Xavier weight distribution can also be used.

\section{Optimization} \label{appendix:optimization}
In the optimization step the parameters $\theta$ of a layer are updated based on the value of their gradient $\Gradient \theta$ with respect to a given loss function. The goal of this step is to decrease the value of the loss. In this section we give three common choices for optimization methods: gradient descent, momentum and Nesterov. All of them take a learning rate parameter $\eta$ as input, which is used to control the size of the optimization step. Our equations are equivalent to the ones in Keras \citep{chollet2015keras}, but presented in matrix-form. 
We use a prime symbol to denote updated values.
\[
\begin{array}{L @{\hspace{20pt}} L @{\hspace{20pt}} L}
\textsc{Gradient descent} & \textsc{Momentum} & \textsc{Nesterov} \\
\addlinespace[1ex]
\theta' = \theta - \eta \cdot \Gradient \theta
&
\begin{aligned}[t]
  \Delta'_\theta &= \mu \cdot \Delta_\theta - \eta \cdot \Gradient \theta \\
  \theta' &= \theta + \Delta'_\theta
\end{aligned}
&
\begin{aligned}[t]
  \Delta'_\theta &= \mu \cdot \Delta_\theta - \eta \cdot \Gradient \theta \\
  \theta' &= \theta + \mu \cdot \Delta'_\theta - \eta \cdot \Gradient \theta
\end{aligned}
\end{array}
\]
Both momentum and Nesterov depend on a parameter $0 < \mu < 1$. Furthermore, they store an additional parameter $\Delta_\theta$ with the same dimensions as $\theta$. This parameter is updated in each optimization call, and initially contains only zeroes.

\section{Learning rate schedulers} 
\label{appendix:learning-rate-schedulers}
We define a learning rate scheduler as a function $\eta: \mathbb{N} \rightarrow \Reals$ that returns the learning rate at optimization step $i$. We assume that $\eta_0$ is a given initial learning rate.
\[
\begin{array}{lll} 
\textsc{Formula} & \textsc{Description} 
\\
  \eta_{i}= \eta_0 
  & \text{ A constant scheduler with initial value $\eta_0$. }
\\
  \eta_{i+1} = \displaystyle \frac {\eta_{i}}{1+d \cdot i} 
  & \text{ A time-based scheduler with decay parameter $d$. } 
\\
\\
  \eta_{i} = \displaystyle \eta_{0} \cdot d^{ \left \lfloor {\frac {1+i}{r}} \right \rfloor } 
  & \text{A step-based scheduler with change rate $d$ and drop rate $r$. } 
\\
\addlinespace[1ex]
  \eta_{i} = \eta_{0} e^{-d \cdot i} 
  & \text{An exponential scheduler with decay parameter $d$.}
\\   
\addlinespace[1ex]
  \eta_{i} = \displaystyle \eta_{0} \Gamma^{\sum_{j=1}^k \lfloor \frac{i}{m_j} \rfloor}
  & \text{A multi-step scheduler with decay parameter $\Gamma$ and milestones $\set{m_1, \ldots m_k}$.}
\end{array}
\]

\end{document}